\documentclass[acmlarge,nonacm]{acmart}

\AtBeginDocument{%
  \providecommand\BibTeX{{%
    Bib\TeX}}}

\usepackage{amsmath,amsfonts}
\usepackage{algorithmic}

\usepackage{amssymb}
\usepackage{graphicx}
\usepackage{threeparttable}
\usepackage{textcomp}
\usepackage{svg}
\usepackage{tabularx}
\colorlet{shadecolor}{gray!40}
\usepackage{tcolorbox}
\usepackage{xcolor}  
\definecolor{highlightblue}{RGB}{205,225,245}
\definecolor{highlightorange}{RGB}{255,224,189}
\usepackage{colortbl}   
\usepackage{float}
\usepackage{graphicx}
\usepackage{booktabs}
\usepackage{subcaption}
\usepackage{geometry}
\usepackage{multirow}
\usepackage{pifont} 
\usepackage{array}
\usepackage{array}     
\usepackage{geometry} 
\usepackage{multirow}
\usepackage{tabularx}
\usepackage{graphicx}
\usepackage{booktabs}
\usepackage{longtable}
\usepackage{ragged2e}
\usepackage{listings}
\usepackage{float} 

\setcopyright{none}

\newcommand{\cmark}{\textcolor{green!60!black}{\ding{51}}}  
\newcommand{\xmark}{\textcolor{red}{\ding{55}}} 

\usepackage{bm}
\def\BibTeX{{\rm B\kern-.05em{\sc i\kern-.025em b}\kern-.08em
    T\kern-.1667em\lower.7ex\hbox{E}\kern-.125emX}}
\renewcommand\footnotetextcopyrightpermission[1]{}

\title[Large Language Models for Oral History Understanding]{Large Language Models for Oral History Understanding with Text Classification and Sentiment Analysis}

\author{Komala Subramanyam Cherukuri}
\email{KomalaSubramanyamCherukuri@my.unt.edu}
\orcid{0009-0009-7474-1390}
\author{Pranav Abishai Moses}
\email{pranavabishaimoses@my.unt.edu}
\orcid{0009-0002-9316-6116}
\author{Aisa Sakata}
\email{AisaSakata@my.unt.edu}
\orcid{0009-0001-8128-0598}
\affiliation{%
  \institution{University of North Texas}
  \streetaddress{3940 N Elm St}
  \city{Denton, TX}
  \country{USA}
}

\author{Jiangping Chen}
\orcid{0000-0002-7016-4583}
\affiliation{%
  \institution{University of Illinois Urbana-Champaign}
  \streetaddress{Library and Information Science Building, 501 E Daniel St}
  \city{Champaign, IL}
  \country{USA}}
\email{jpchen@illinois.edu}

\author{Haihua Chen}
\orcid{0000-0002-7088-9752}
\authornote{Corresponding author.}
\affiliation{%
  \institution{University of North Texas}
  \streetaddress{3940 N Elm St}
  \city{Denton, TX}
  \country{USA}}
\email{haihua.chen@unt.edu}

\renewcommand{\shortauthors}{Cherukuri, et al.}
\begin{document}

\begin{abstract}

Oral histories serve as vital records of lived experience, particularly within communities affected by systemic injustice and historical erasure. Effective and efficient analysis of their oral history archives can promote access and understanding of the oral histories. However, large scale analysis of these archives remains limited due to their unstructured format, emotional complexity, and the high cost of manual annotation. This paper aims to develop a scalable framework to automate semantic and sentiment annotation for oral history archives, with a particular focus on Japanese American Incarceration Oral History (JAIOH). Using large language models (LLMs), this study seeks to construct a high-quality dataset, systematically evaluate the performance of multiple LLMs, and investigate effective prompt engineering strategies on annotation in historically sensitive contexts. Our multiphase approach combines expert annotation, prompt design, and LLM evaluation using ChatGPT, Llama, and Qwen. We labeled 558 sentences from 15 narrators for sentiment and semantic classification \footnote{The third author of this paper is an domain expert.}, then developed prompts and evaluated across zero shot, few shot, and retrieval augmented generation (RAG) strategies based on the labeled data. For semantic classification, ChatGPT achieved the highest F-1 score (88.71\%), followed by Llama (84.99\%) and Qwen (83.72\%). Regarding sentiment analysis, Llama performed slightly better (82.87\%) than Qwen (82.66\%) and ChatGPT (82.29\%), with all models showing comparable results. Based on the overall evaluation, we selected the best performing prompt configurations for each task and used them to automatically annotate 92,191 sentences from 1,002 interviews in the JAIOH collection. This paper introduces a large scale annotated oral history corpus and offers a comparative benchmark for prompt based LLM annotation in low resource, culturally sensitive domains. Our findings demonstrate that LLMs can effectively perform semantic and sentiment annotation across large oral history collections when guided by carefully designed prompts. This study contributes both a reusable annotation pipeline and practical guidance for applying LLMs in culturally sensitive archival analysis. By bridging archival ethics with scalable NLP techniques, this work lays the groundwork for responsible use of artificial intelligence in digital humanities and preservation of collective memory. All code, annotated data, prompt templates, and experimental details are available at the \href{https://github.com/kc6699c/LLM4OralHistoryAnalysis}{Repository}.

\end{abstract}






\maketitle

\section{Introduction}


Oral history preserves the lived experiences and cultural memory often absent from formal archives, offering a counter-narrative to institutional records. Long before the development of writing systems, spoken narratives served as the primary means through which communities transmitted knowledge, cultural values, and collective identity across generations. This tradition remains significant today, particularly in societies where written documentation is limited or inaccessible. Oral accounts capture personal perspectives, emotional nuance, and cultural knowledge that rarely appear in official records. By amplifying the voices of those historically excluded from dominant narratives, oral history contributes to a more inclusive and multidimensional understanding of the past. For example, testimonies of Japanese American incarceration survivors, particularly those presented during the Commission on Wartime Relocation and Internment of Civilians, played a key role in shaping public memory and catalyzing the redress movement, ultimately contributing to the passage of the Civil Liberties Act of 1988 \cite{nagata2015processing}. These narratives reveal aspects of daily life, cultural continuity, and memory that formal records often overlook. The loss of such accounts would create substantial gaps in the way history is understood, especially in relation to identity and collective experience \cite{jones2021voice}. Oral history serves both as a method and a cultural practice to deepen historical records. Advances in digital technology have enabled these archives to be digitized and made widely accessible. These collections capture personal stories, cultural memories, and community experiences. However, these archives often contain unstructured content, such as long-form audio recordings or transcripts, which pose challenges for search, analysis, and interpretation using conventional methods. As a result, researchers have traditionally relied on manual annotation to extract meaning and identify patterns within these narratives \cite{sun2024application}. Addressing these challenges is essential for unlocking the full research potential of oral history in the digital age.

Despite their value, oral histories remain difficult to study on a scale. They are deeply personal, emotionally rich, and shaped by context, making it difficult to apply one-size-fits-all analysis techniques. Researchers have emphasized the need for rigorous methodological guidelines when working with oral history data, particularly to avoid misrepresenting nuanced personal testimony or applying reductive classifications to complex narratives. The interpretive ambiguity, cultural specificity, and emotional depth inherent in these accounts make them difficult to process using standardized computational techniques. As a result, oral histories remain underutilized in large-scale research, despite the growing availability of digitized archives \cite{firouzkouhi2015data}.

Natural language processing (NLP) offers a promising way to overcome these barriers. NLP techniques have been applied to Densho oral histories \cite{chen2024demystifying}, including spaCy phrase matching to extract demographic metadata, KeyBERT for unsupervised keyword extraction, BERTopic to group semantically coherent categories, rule-based sentiment analysis using predefined polarity lexicons, and SBERT-based similarity to categorize narrative phases. With NLP, researchers can automatically identify classes, emotions, and patterns in large collections of oral history transcripts. These tools can also help detect subtle elements, such as mood, cultural expression, or emotional intensity insights that would be difficult to uncover by reading or listening manually. This is especially useful when analyzing thousands of interviews, which would be too time consuming for humans alone \cite{pessanha2021computational}. 

New developments in NLP and text mining make it easier to apply these methods to historical research. While many of these tools were first built for modern, well-structured data, researchers are now adapting them for oral histories and other historical texts. Advances in NLP and large language models (LLMs) have opened new possibilities for analyzing unstructured historical data. NLP systems, LLMs such as ChatGPT, Llama, and Qwen can perform context sensitive classification and summarization tasks with minimal supervised training, allowing scalable analysis of emotionally and culturally nuanced oral histories \cite{li2024evaluating}. LLMs are capable of performing tasks through zero-shot and few-shot in-context learning, achieving competitive results without requiring domain-specific supervised training \cite{zhang2023sentiment}. These methods allow scholars to explore long interviews more efficiently, helping to uncover deeper insights about the past through personal stories \cite{brown2023text}. 

To ground our approach in a culturally rich and historically underrepresented corpus, we selected the Densho Digital Repository's Japanese American Incarceration Oral History (JAIOH) collection. Previous work has demonstrated the importance of building fine-grained annotation schemes to uncover marginalized narratives and allow large-scale processing of this corpus through NL techniques \cite{han2022uncover}. Manually analyzing thousands of oral history interviews poses challenges due to the subtle, layered, and often ambiguous nature of spoken narratives. This complexity is particularly pronounced in Japanese-American oral histories, where emotional expression may be subdued and meaning is frequently conveyed through implicit or culturally nuanced language. Such narratives resist straightforward interpretation, making it difficult to scale analysis while preserving the authenticity and depth of lived experience. To enable a rigorous and scalable analysis of these materials, a high-quality labeled data set is essential.

We explored the GPT-4o (ChatGPT), Qwen-2.5-32B and Llama-3.3-70B-Versatile, to perform semantic and sentiment analysis on oral history transcripts related to Japanese-American incarceration. In the initial phase, we manually annotated interview transcripts from a representative sample of 15 narrators, producing sentence-level labels for both thematic categories and sentiment. These annotations served as a benchmark for evaluating model performance. We first evaluated LLM output against human-labeled samples on a small corpus of 558 sentences to identify the most effective prompt strategy. In the second phase, we focused on optimizing the behavior of LLMs through prompt engineering. We designed and tested four prompt variants for sentiment and five for semantic classification across four different strategies to identify which formulations yielded the most accurate and contextually appropriate output when applied to our labeled sample. Based on this evaluation, we selected the most effective prompt configuration for each model and used it to automatically label the entire corpus, which includes transcripts from over 1,000 interviews, followed by downstream analysis using topic modeling (BERTopic) and entity extraction to assess semantic consistency and narrative coverage. By combining manual annotation with prompt-based LLM inference, this work enables a large-scale, nuanced analysis of historically significant narratives. The resulting annotated corpus supports a deeper engagement with marginalized voices, while the methodology offers a transferable framework for applying LLMs to other oral history archives.
Through the implementation of these methodologies, this study seeks to address the following questions:
\begin{enumerate}
    \item How can we create a high-quality large-scale dataset to automatically understand Japanese-American oral history?
    \item What factors should be considered when selecting the most effective and efficient prompt strategy for annotating historically sensitive oral history data using LLMs?
    \item To what extent can LLMs preserve contextual nuance and narrative integrity when applied to large scale annotations of testimonies from marginalized communities?
    \item What are the limitations and strengths of LLM-driven annotation pipelines in producing reliable and scalable semantic and sentiment labels for under-described oral history collections?
\end{enumerate}
By exploring these questions, we aim to showcase the potential of advanced language models in enhancing the interpretation and analysis of complex historical narratives.

\section{Related Work}

In this section, to summarize the existing efforts and highlight current gaps, we review literature related to our paper from three aspects: (1) Existing efforts in understanding different kinds of oral history, including Japanese American Incarceration Oral History, (2) the applications of artificial intelligence (AI), machine learning (ML), natural language processing (NLP), and other techniques for understanding oral history archival collections, (3) large language models (LLMs) or generative AI (GAI) for text analysis, especial automatic text classification and sentiment analysis with a focus on digital archives.

\subsection{Existing Efforts in Understanding Oral History Interpretation}

The digitization of historical records has long been a concern in the archival and digital humanities communities. Existing efforts have focused on transforming paper-based archives into machine-readable formats to enable information retrieval, mining, and preservation on a scale \cite{girdhar2024digitizing}, and oral history in community learning \cite{mclellan1998case}. These works emphasize the technical infrastructure for digital conversion \cite{tsuchiya2024implementation}, but also the interpretive potential of computational tools to uncover historical narratives \cite{heckman2023shifting}. Parallel to these digitization efforts, oral histories have been studied for their unique value in capturing personal, community, and marginalized voices that are often absent from official records \cite{swain2003oral, alexander2006excluding}. Scholars have investigated the methodological frameworks \cite{stephan2021platinum} and ethical imperatives \cite{thompson2015voice, flinn2009whose} of collecting, curating, and describing oral testimonies, emphasizing inclusivity, authenticity, and sensitivity to archival silences. Practical guidance on the curating of oral histories has also been provided in archival contexts \cite{mackay2016curating}.

In particular, Japanese-American incarceration during World War II has become a critical case study in oral history and trauma archiving. Studies have examined systemic violence and racial trauma experienced by Japanese Americans \cite{laher2006internment, nagata1998coping, nagata2019japanese}, alongside community-driven efforts to preserve memories \cite{odo2017memorializing, ikeda2014densho} and official archival documentation \cite{nationalarchives_japanese_incarceration}. These narratives have been foundational in shaping public memory and academic discourse around national trauma, race, and historical justice.

\subsection{AI/ML/NLP in Understanding Oral History Archival Collections}
Recent computational approaches have introduced multimodal techniques for analyzing oral history archives, employing automatic speech recognition (ASR), transformer based language models, and social signal processing to handle multilingual and emotionally charged interviews \cite{sabra2025deciphering}. Oral history archives have also been examined from a computational perspective \cite{pessanha2021computational}. Studies have explored the challenges of emotion annotation in oral histories and modality dependent inconsistencies \cite{gref2022study, de2013emotional}.
For thematic analysis, n-gram statistics, topic modeling, and temporal lexical tracking have been used to extract sociocultural patterns and diachronic ideological shifts \cite{brown2023text, pawlowski2024nlp}. NLP tools originally developed for structured or literary corpora such as lemmatizers, part-of-speech taggers, and named entity recognition (NER) systems have been repurposed for oral history contexts \cite{brooke2015gutentag}. Studies have examined how linguistic cues such as hedging and implicature contribute to the construction of epistemic stance in politically sensitive oral histories \cite{fitzgerald2020penetrating}.
Structural modeling of ASR transcripts has been explored to improve segmentation and alignment with narrative structure \cite{olsson2007improving}. Oral histories have also been studied as historiographic sources for recovering underrepresented perspectives, including in computing history \cite{nyhan2015oral}.
Multimodal archival systems now integrate ASR, facial recognition, and topic modeling to support scalable content retrieval and interactive exploration \cite{sichani2021connected}. Broader machine learning frameworks for cultural heritage applications are described in recent surveys \cite{fiorucci2020machine}, while best practices for responsible deployment of ML in archives are addressed through operational checklists \cite{lee2025collections}.
Conversational systems and retrieval models tailored to archival corpora have also emerged. These include question generation models using BART with semantic constraints for French-language archives \cite{bechet2022question}, and agents like CulturalERICA designed to facilitate the exploration of collections of European cultural heritage \cite{machidon2020culturalerica}. Complementary work addresses machine learning methods for processing spoken audio in libraries, archives, and museums (LAM) \cite{xu2020study}, and resources such as the Czech Historical Named Entity Corpus support NER and linguistic annotation for historical corpora \cite{hubkova2020czech}.

\subsection{Large Language Models/Generative AI for Digital Archives}
LLMs have recently been applied to oral history archives for tasks such as sentiment analysis, classification, and contextual annotation \cite{ashqar2025sentiment, guo2024prompting, aguiar2024final}, model adaptation for historical texts \cite{manjavacas2022adapting}. These approaches reflect a growing interest in using generative AI to support interpretive and analytical processes in the digital humanities. Emerging work also addresses responsible and ethical deployment of AI in archival contexts, with a focus on trust, transparency, and collaborative institutional practices \cite{mannheimer2024responsible, jaillant2023applying}.
Specialized corpora are developed to support large-scale NLP tasks in historical domains, including sentiment classification, named entity recognition, and thematic modeling \cite{jaff2025corhoh}. RAG has been explored to improve dialogue modeling, query reconstruction, and context tracking in testimonial archives \cite{zhang2025dh, karimzadeh2025reconstructing}. Knowledge graph construction frameworks using LLM-RAG have also been proposed to enable structured exploration of oral history content \cite{sun2024application}.
Multimodal and generative models are increasingly used for transcription and metadata generation. Tools such as Transcription Pearl apply LLMs to mass transcription of handwritten archival material \cite{humphries2025unlocking}. Related initiatives have developed question-answering datasets for historical newspapers \cite{piryani2024chroniclingamericaqa}, and AI tools for linking community-contributed cultural heritage collections \cite{hannaford2024our}.

While prior works explored oral history archiving, trauma representation, and computational analysis using NLP and LLM. Existing research often focuses on pipeline development or case-specific applications, lacking rigorous benchmarking or prompt evaluation in historically sensitive contexts. This gap is particularly significant for underrepresented collections such as Japanese-American incarceration testimonies, where precision, nuance, and contextual integrity are critical. This study extends our earlier corpus-level analysis \cite{chen2024demystifying}, our approach offers a framework for automating sentiment and semantic classification in incarceration-era Japanese American oral histories. In this work we address this by building a labeled corpus and evaluating multiple LLM prompting strategies, offering insights into model behavior, annotation reliability, and narrative preservation at scale.

\section{Methodology}

\subsection{Research Design}

\begin{figure}[H]
    \centering
    \includegraphics[width=1.0\textwidth,height=0.28\textheight]{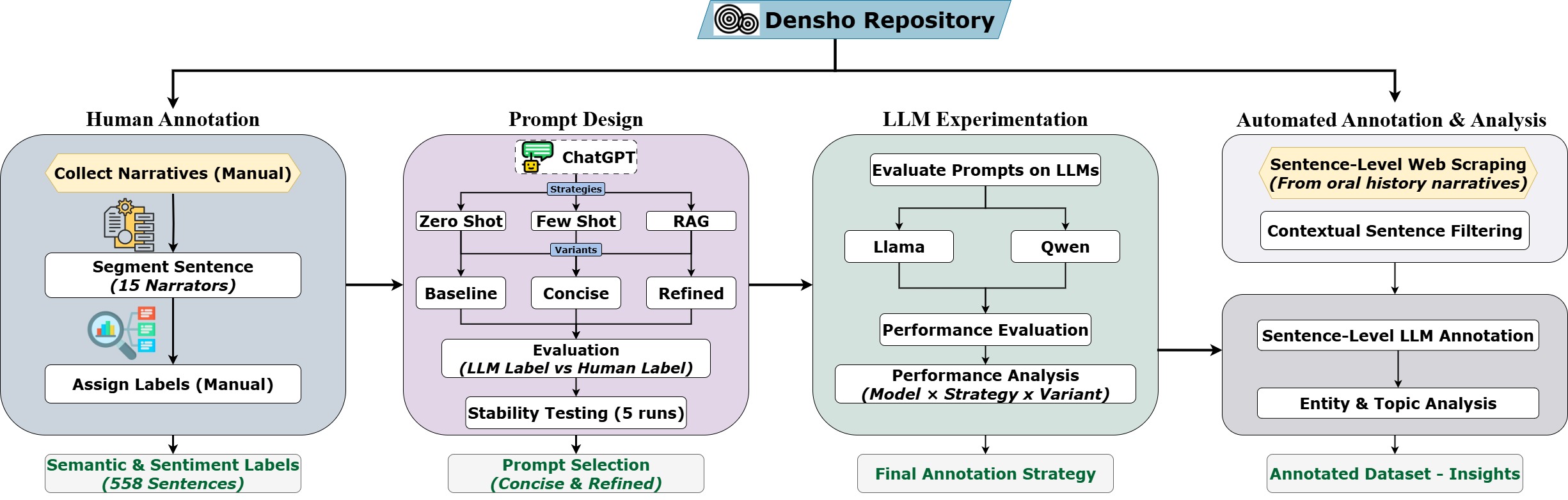}
    \Description{A diagram showing the annotation and evaluation pipeline used to process Japanese-American incarceration oral histories using large language models. The flow includes data preprocessing, model prompting, label generation, and evaluation.}
    \caption{Oral History Text Annotation and Evaluation Using LLMs}
    \label{fig:RD}
\end{figure}

Our study proposes a framework for sentence-level classification and sentiment analysis of oral histories on Japanese-American incarceration from the Densho repository. The framework integrates expert human annotation, prompt engineering, and large language model (LLM) evaluation to address the complexities of culturally sensitive narratives. Figure~\ref{fig:RD} presents an overview of our annotation and evaluation pipeline. The framework consists of four phases: Human Annotation, LLM Experimentation, LLM Evaluation \& Selection, and Automated Annotation \& Analysis. Each phase produces an output, shown in green, which serves as the input for the next phase. Our research design demonstrates the feasibility of combining expert annotation with LLMs for structured, scalable analysis. This approach supports digital history archiving and establishes a culturally informed and methodologically rigorous foundation for future computational research in the digital humanities.

\subsection{Data Collection and Preprocessing}

This study leverages a curated dataset derived from the Densho Digital Repository\footnote{\url{https://densho.org/}}, a comprehensive archive dedicated to preserving Japanese-American oral histories. The Densho repository was selected due to its integration of multiple partner collections, including those of the Japanese American Service Committee (JASC), the Japanese American Museum of San Jose (JAMSJ), and Discover Nikkei, which offer centralized access to a wide array of historically significant interviews and transcripts related to incarceration and postwar experiences.

The complete data set includes 1002 oral history transcripts from the narrators. These transcripts were programmatically parsed using Python, and the data was extracted at the sentence level. A key challenge in processing this content arises from its heterogeneous structure: transcripts vary considerably in length and are typically formatted in a question-and-answer style. To address this, only the responses from the narrators were retained, filtering out interviewer prompts to focus the analysis on personal narrative content. For the initial phase of annotation and prompt development, a representative sample comprising 558 sentences was selected from 15 narrators. These sentences were manually extracted by human readers to ensure contextual and interpretive fidelity. A two-stage annotation process was used: in the first stage, an annotator curated sentence-level segments for clarity and relevance; in the second stage, a domain expert in Japanese-American incarceration history provided annotations for both classification and sentiment labels. This protocol ensured both linguistic consistency and historical accuracy in the resulting labeled dataset, establishing a robust foundation for subsequent natural language processing (NLP) tasks.

\subsection{Data Annotation}


\subsubsection{Annotation Schemes}\hfill\\
We adopted a dual layer annotation scheme which contains sentiment and semantic labels. The sentiment labels include Positive, Neutral, and Negative,  are the commonly used sentiment labels for sentiment analysis \cite{gref2022study}. The semantic labels are six categories, including : Biographical Information, Life Before Removal, Life During Incarceration, Military Service, Returning of Japanese Americans After WWII, and Movements of Peace and Justice. These categories were adapted from our previous work \cite{han2022uncover}, which was empirically derived from the large scale analysis of Densho interviews and based on established historiographic sources. This scheme is able to capture both the emotional tone and historical relevance of each sentence, enabling a structured and nuanced understanding the oral history content.


\subsubsection{Sentiment Labeling}\hfill\\
To capture the emotional tone embedded in historically rich narratives, each sentence was manually labeled Positive, Neutral, or Negative, reflecting implicit and context dependent sentiment. A native Japanese speaker performed the initial labeling to ensure cultural and linguistic accuracy, followed by independent verification by an additional member of the team. This rigorous two stage process was essential to ensure consistency and reliability, particularly given the nuanced and sensitive nature of the content.

\subsubsection{Semantic Labeling}\hfill\\
As presented in Table~\ref{tab:Description_classes}, to enable structured analysis of Japanese American oral histories, we adopted a semantic annotation scheme comprising six predefined categories, adapted from our previous work \cite{han2022uncover}, which was empirically derived from the large-scale analysis of Densho interviews and based on established historiographic sources.

\begin{table}[htbp]
\vspace{-0.3em}
\centering
\renewcommand{\arraystretch}{1.4}
\begin{tabular}{
    >{\bfseries}m{1.3cm} 
    >{\RaggedRight\arraybackslash}p{3.5cm} 
    >{\RaggedRight\arraybackslash}p{10.5cm}
}
\toprule
\textbf{Index} & \textbf{Semantic Class Label} & \textbf{Description} \\
\midrule
0 & Biographic Information & Captures foundational details about the narrator, including names, birth stories, family history (limited to birthdays and migration to the US), and places of residence. \\
1 & Life Before Removal & Contextualizes the buildup to internment, including anti-Japanese sentiment, discriminatory legislation, economic and social challenges, and the broader political climate that precipitated forced removal. \\
2 & Life During the Incarceration & Details about life inside internment camps, including living conditions, cultural adaptation, personal accounts, and social dynamics under confinement. \\
3 & Military Services & Highlights the wartime contributions of Japanese Americans, with emphasis on the 100th Battalion and the 442nd Regimental Combat Team, units noted for valor and distinction despite prevailing discrimination. \\
4 & Returning of Japanese Americans After WWII & Documents the return and reintegration process following incarceration, including migration patterns, community rebuilding, and the socioeconomic obstacles faced by returnees. \\
5 & Movements for Peace and Justice & Focuses on post-incarceration activism for redress, civil rights restoration, and reparations, particularly during the mid-to-late 1980s. It also includes aspirational expressions for future peace and justice. \\
\bottomrule
\noalign{\vskip 5pt}
\end{tabular}
\caption{Semantic Class Definitions for Oral History Narratives}
\label{tab:Description_classes}
\vspace{-1.5em}
\end{table}

This classification framework was applied to all 558 sentence level entries, with manual annotation ensuring high fidelity to narrative context and thematic intent. This process serves as a foundational step in assessing the viability of NLP techniques to process culturally embedded historical narratives.

\subsubsection{Human Annotation}\hfill\\
The annotation process was conducted in two distinct phases. In the first phase, a researcher manually identified contextually rich and semantically complete sentences suitable for classification. In the second phase, a domain expert, fluent in Japanese and deeply familiar with the cultural and historical nuances of incarceration narratives, labeled each sentence with both a semantic category and a sentiment label. This expert-led approach ensured that subtle emotional signals and historically embedded meanings were accurately captured. The final data set comprises six semantic categories (classes 0 to 5), distributed as follows: class 0 (Biographical Information) with 30 sentences, class 1 (Life before removal and during incarceration) with 111 sentences, class 2 (Life during incarceration) with 253 sentences, class 3 (military service) with 72 sentences, class 4 (Returning after WWII) with 61 sentences, and class 5 (Movements for peace and justice) with 31 sentences. In terms of sentiment, 289 sentences were labeled as neutral, 98 as positive, and 171 as negative.

To further ensure consistency and reliability, both annotators independently verified each label. Any discrepancies were discussed and resolved through collaborative review, resulting in a consensus annotation. This rigorous process lasted five weeks, reflecting the care required to accurately curate historically sensitive oral histories. We note that conventional metrics of agreement between annotators (for example, Cohen's Kappa) were not calculated because our annotation followed a consensus based protocol rather than independent parallel labeling. Sentence extraction and labeling were treated as separate complementary tasks performed by two different annotators. The final labels were established through expert validation and joint resolution of disagreements, ensuring historical fidelity and interpretive precision. This approach aligns with established practices in digital humanities and oral history research, where annotation quality and cultural sensitivity are prioritized over raw statistical agreement.

\subsection{LLM Experimentation}

Prompts are structured inputs designed to guide LLMs toward producing outputs that conform to specific goals, formats, or constraints. Serving as a form of high-level programming, prompts allow users to control model behavior, automate tasks, and align outputs with defined objectives \cite{white2023prompt}. The process of constructing these prompts, known as prompt engineering, involves deliberate and iterative refinement of instructions to improve clarity, interpretability, and alignment of tasks \cite{ye2023prompt}. This methodology is particularly critical in settings where generic instructions lead to ambiguous or low-quality results and where nuanced control over model reasoning is required \cite{dang2022prompt}.
In this study, we employed a combination of zero-shot, few-shot, and retrieval-augmented prompting techniques to support the dual annotation tasks of Semantic and sentiment analysis. Each strategy was selected based on its ability to align with the underlying reasoning demands of the task and the nature of the oral history data.

In this study, we used a combination of zero-shot, few-shot, and retrieval-augmented prompting strategies to support dual annotation tasks of semantic and sentiment analysis. Each approach was chosen based on its ability to address the interpretive demands of the task and the linguistic complexity of the oral history transcripts. A brief overview of these prompting strategies is provided below:

\begin{itemize}
    \item \textbf{Zero Shot:} In the zero shot setting, the model is prompted with a simple instruction or question, without examples, relying entirely on its pretrained knowledge to infer the task and generate a response \cite{kojima2022large}. Although this method is efficient and does not require manual data preparation, its success is highly sensitive to prompt.
    \item \textbf{Few Shot:} Few-shot prompting improves upon this by embedding a small set of labeled examples directly into the prompt. These demonstrations serve as reference points for the model, helping it generalize to new and unseen instances \cite{chen2024prompts, li2023practical}. This format is especially useful when the task involves domain-specific output patterns or subtle distinctions that are not easily inferred from instructions alone.
    \item \textbf{Retrieval Augmented Generation:} To further enhance contextual accuracy, we also employed a RAG strategy. RAG augments the prompt with dynamically retrieved context-relevant content from external sources, providing the model with additional background information to improve factuality, disambiguation, and semantic grounding \cite{wu2024clasheval}. This was particularly valuable when the input text lacked sufficient local cues to determine the class or emotional tone reliably.
\end{itemize}

Together, these prompting strategies allowed us to tailor model behavior to the specific goals of each annotation task. The task specific prompts consistently outperform generic ones, although their development requires considerable human effort \cite{reynolds2021prompt}. Although both semantic and sentiment analysis are supervised classification tasks, they involve distinct cognitive and linguistic cues. Semantic analysis requires identification of the semantic domain or subject matter (for example, 'biographical information', 'military service'), while sentiment analysis focuses on detecting the evaluative tone (for example, 'positive', 'negative'). Designing a single prompt to perform both tasks risks reducing the interpretive accuracy for one or both objectives.
Moreover, recent work demonstrates that the structure of prompts strongly influences the behavior of LLM reasoning. For example, chain-of-thought prompting has been shown to significantly enhance performance by encouraging intermediate reasoning steps \cite{wei2022chain}. These findings underscore the importance of crafting task-specific prompts that explicitly model the reasoning pathways relevant to each annotation goal. In our workflow, we therefore adopted differentiated prompt strategies for semantic and sentiment classification, ensuring that the interpretive focus of the model aligned with the unique requirements of each task.

\subsection{Prompt Design}
The components in a prompt are crucial because they define the structure, context, and clarity needed to guide the behavior of a language model. By thoughtfully constructing these components, we can more effectively communicate our intention, resulting in more accurate, relevant, and meaningful outputs aligned with our goals \cite{giray2023prompt}. The sentiment and semantic classification prompt templates are provided in Appendix~\ref{sec:prompt_templates}, with full prompts on GitHub.

\begin{table}[htbp]
\centering
\begin{tabular}{>{\centering\arraybackslash}p{0.6cm}|>{\raggedright\arraybackslash}p{4.69cm}|c|c|c|c|c}
\hline
\textbf{Task} & \textbf{Component} & \textbf{Foundational} & \textbf{Structured} & \textbf{Comprehensive} & \textbf{Refined} & \textbf{Concise} \\
\hline
\multirow{5}{*}{\rotatebox[origin=c]{90}{\textbf{Sentiment}}} 
& Instruction           & \cmark & \cmark & \cmark & --     & \cmark \\
& Sentiment List        & \cmark & \cmark & \cmark & --     & \xmark \\
& Sentiment Definition \& Examples & \xmark & \cmark & \cmark & --     & \xmark \\
& Background            & \xmark & \xmark & \cmark & --     & \xmark \\
& Concise Summary       & \xmark & \xmark & \xmark & --     & \cmark \\
\hline
\multirow{7}{*}{\rotatebox[origin=c]{90}{\textbf{Semantic}}} &
Instruction           & \cmark & \cmark & \cmark & \cmark & \cmark \\
& Category List         & \cmark & \cmark & \cmark & \cmark & \xmark \\
& Category Definitions  & \xmark & \cmark & \cmark & \xmark & \xmark \\
& Background            & \xmark & \xmark & \cmark & \cmark & \xmark \\
& Keywords              & \xmark & \xmark & \cmark & \cmark & \xmark \\
& ChatGPT Definitions   & \xmark & \xmark & \xmark & \cmark & \xmark \\
& Concise Summary       & \xmark & \xmark & \xmark & \xmark & \cmark \\
\hline
\noalign{\vskip 5pt}
\end{tabular}
\caption{Comparison of Prompt Components Used Across Two Classification Tasks: Sentiment and Semantic}
\label{tab:combined_tasks}
\vspace{-1.65em}
\end{table}

\subsubsection{Sentiment Classification}\hfill\\
For the sentiment classification task, we developed prompt based methods to label each sentence as Positive, Neutral, or Negative. The prompts were designed with clear task instructions, sentiment definitions, and illustrative keywords to minimize ambiguity and guide model predictions. Table~\ref{tab:componentsentiment} outlines the key components used in our sentiment prompt design.

\begin{table}[htbp]
\vspace{-0.3em}
\centering
\renewcommand{\arraystretch}{1.4}
\begin{tabular}{>{\bfseries} p{3.9cm} p{12.1cm}}
\toprule
\textbf{Prompt Component} & \textbf{Description} \\
\midrule
Instruction & This component defines the structure of the task by specifying the input type (sentence) and the expected output (label: \textit{positive, negative}, or \textit{neutral}). \\
Sentiment List & The sentiment classification task uses a set of fixed labels consisting of three categories: \textit{Positive}, \textit{Negative}, and \textit{Neutral}. Each label represents the emotional tone expressed in the input text. \\
Sentiment Definition \& Examples & Each sentiment category was defined, and with minimal examples with respect to the historical and emotional context of Japanese American oral histories. These definitions reflect emotional expression in personal recollections of incarceration. \\
Concise Summary & A concise prompt was generated by combining the contextual background with simplified sentiment definitions. This summary offers a clear and concise version of the task, helping the model understand and perform sentiment classification accurately and consistently. \\
\bottomrule
\noalign{\vskip 5pt}
\end{tabular}
\caption{Components of the Sentiment Classification Prompt for Japanese American Oral Histories}
\label{tab:componentsentiment}
\vspace{-1.65em}
\end{table}

To support high-quality and context aware sentiment annotation, we developed prompt variants with increasing levels of guidance. Table~\ref{tab:combined_tasks} outlines their structure, this progression reflects our design strategy: to balance contextual informativeness with prompt efficiency. From minimal instructions in the foundational prompt to a concise and historically grounded linguistically streamlined version, each variant was designed to improve interpretive precision in emotionally nuanced incarceration narratives. The concise prompt, in particular, was optimized using ChatGPT to retain cultural and emotional specificity while improving clarity and scalability for large-scale annotation.




\subsubsection{Semantic Classification}\hfill\\
To implement semantic classification effectively, we decomposed the prompt into a set of modular components, each serving a distinct purpose in guiding model interpretation and output consistency. Below, we describe these components in detail.

\begin{table}[htbp]
\centering
\renewcommand{\arraystretch}{1.4}
\begin{tabular}{>{\bfseries} p{3.5cm} p{12cm}}
\toprule
\textbf{Prompt Component} & \textbf{Description} \\
\midrule

Instructions & This component defines the task structure by specifying the input type (sentence-level transcript) and the expected output (a class or category number). It ensures that the classification process remains focused and consistent, particularly when dealing with historically grounded oral narratives. \\
Category List & This component provides a numbered list of categories, each representing a distinct theme. These numbers are used throughout the classification process to maintain an organized and consistent labeling system. \\
Definition of Each Category & This component offers clear explanations of each category including, grounded in historical context. It helps distinguish between different types of content and ensures that each category is applied appropriately. \\
Background & This component provides an essential historical context for the incarceration of Japanese Americans during World War II. By briefly summarizing key events and sociopolitical dynamics, it helps to ground the classification task in its broader ethical and historical framework. \\
Keywords & This component includes commonly associated words for each category. These terms guide annotators and models in identifying relevant class in the transcripts and support more focused and efficient classification. \\
ChatGPT Definition & The original definitions and background were rephrased using ChatGPT to improve clarity and focus while keeping the original content. The refined version was verified by a human expert to confirm their accuracy and usefulness for annotation. \\
Concise Summary & Original ChatGPT category definitions, key words, and background descriptions were refined using ChatGPT to improve linguistic clarity, consistency, and instructional precision. This process preserved the core historical meanings while improving the readability of the model. Each refined definition was subsequently reviewed and validated by a domain expert to ensure interpretive fidelity and practical utility in annotation workflows. \\
\bottomrule
\noalign{\vskip 5pt}
\end{tabular}
\caption{Components of the Semantic Classification Prompt for Japanese American Oral Histories}
\label{tab:semanticomponents}
\vspace{-1.5em}
\end{table}

To systematically improve the accuracy and consistency of LLM-based classification and annotation, we developed a series of prompt variants with increasing levels of guidance, components are shown in Table~\ref{tab:semanticomponents} and components used in each prompt variant shown in Table~\ref{tab:combined_tasks}. To improve LLM performance on the culturally sensitive classification task, we systematically improved prompt design to address ambiguity, enhance contextual understanding, and ensure consistent interpretation. Each successive variant aimed to address specific limitations of the previous one, clarifying category definitions, introducing historical context, and redefining the language for precision. The choice of prompt depends on balancing clarity, context, and efficiency to support accurate and scalable annotation.


\subsection{Experiment Settings}

We evaluated three large language models: GPT-4o (OpenAI), Qwen-2.5-32B, and Llama-3-70B-Versatile (both accessed via Grok API)—using remote API endpoints executed from a Google Colab environment. All inferences were performed via external APIs and no local GPUs were used. Since API usage incurred cost, prompt length and execution efficiency were tightly controlled. GPT-4o was accessed through OpenAI's chat / completion endpoint with temperature = 0 to ensure deterministic results. Qwen-2.5-32B and Llama-3-70B were accessed using Grok-specific APIs, which required model-specific schema formatting. In both cases, the prompts followed the instruction-tuning conventions defined by Grok, and the default system role was applied unless explicitly overridden. Zero-shot and few-shot prompts were executed entirely within the Colab notebook by constructing prompts as raw strings and submitting them through synchronous API calls, the RAG setup required additional configuration. A fixed set of relevant examples retrieved offline using sentence embedding similarity was manually uploaded to the Colab environment prior to inference. This RAG context was uniformly appended to the prompt for all test inputs during that run. The combined prompt, consisting of the RAG examples followed by the test sentence, was submitted through the same API pipeline.

\subsection{Evaluation Metrics}



We use the F1 score as our primary evaluation metric, as it effectively balances correct predictions and classification errors in a single value~\cite{riyanto2023comparative}. To ensure robustness, each prompt was executed five times, and we report the average F1 score across runs. F1 Score defined as:

\begin{equation}
\text{F1} = \frac{TP}{TP + \frac{1}{2}(FP + FN)}
\end{equation}

Here, True Positives (TP) are correctly predicted labels, False Positives (FP) are incorrect predictions, and False Negatives (FN) are missed correct labels. This formulation is equivalent to the harmonic mean of precision and recall but provides a more direct view based on prediction counts.

\subsection{Large Scale Corpus Collection and Processing}
We collected a large-scale corpus of oral history narratives by web scraping the Densho Digital Repository, resulting in 1,002 transcripts comprising over 600,000 sentences. The narratives were segmented into sentences and subjected to rigorous processing, including duplicate removal and validation of scraping integrity. To remove irrelevant sentences, we implemented a two-stage filtering process. First, we performed a manual inspection of a representative subset. Then, we applied binary classification using a fine-tuned Llama-based large language model. This classification step was performed three times to ensure consistency and mitigate possible model hallucinations. After filtering, the final curated dataset consisted of 92,191 sentences, representing historically relevant and high-quality data suitable for downstream classification and sentiment analysis tasks.

\section{LLM Evaluation and Selection}

\subsection{Sentiment Analysis}

To extend our prompt engineering framework to a different linguistic task, we applied our method to sentiment analysis using ChatGPT. The task involved classifying sentences from Japanese American oral history transcripts into one of three sentiment categories: \textit{Positive, Negative, or Neutral}. We evaluated four prompt variants—\textit{Foundational, Structured, Comprehensive, and Concise} across four prompt strategies: \textit{zero shot, zero shot RAG, few shot, and few shot RAG}. We report the F1 score for these trials. The structure and components of each prompt are outlined in Table \ref{tab:combined_tasks}, while the performance results are shown in Table \ref{tab:combined_results_sentiment}.

\subsubsection{Evaluation of LLM Performance for Sentiment Analysis}\hfill\\
We begin by presenting the performance results of ChatGPT under four experimental conditions: zero shot, few shot, zero shot, and few shot with retrieval-augmented generation (RAG), using four different prompt strategies.

\begin{table*}[htbp]
\centering
\small
\begin{tabular}{cllcccccc}
\toprule
\multirow{2}{*}{\textbf{Model}} & 
\multicolumn{1}{c}{\multirow{2}{*}{\textbf{Strategy}}} & 
\multicolumn{1}{c}{\multirow{2}{*}{\textbf{Variant}}} & 
\multicolumn{5}{c}{\textbf{Stability Test}} & 
\multirow{1}{*}{\textbf{Average /}} \\
\cline{4-8}
 & & & \textbf{Test 1} & \textbf{Test 2} & \textbf{Test 3} & \textbf{Test 4} & \textbf{Test 5} & \textbf{Standard Deviation [SD]}\\
\hline
\multirow{16}{*}{\rotatebox{90}{\textbf{\textit{ChatGPT}}}}
& \multirow{4}{*}{Zero-shot} 
    & Foundational & 79.75\% & 78.14\% & 78.85\% & 80.11\% & 80.29\% & 79.43\% [SD= 0.90] \\
    & & Structured & 77.24\% & 77.96\% & 78.32\% & 76.34\% & 76.32\% & 77.24\% [SD= 0.91] \\
    & & Comprehensive & 75.45\% & 75.99\% & 76.11\% & 75.09\% & 74.55\% & 75.44\% [SD= 0.42] \\
    & & Concise  & 81.89\% & 82.25\% & 81.89\% & 82.61\% & 82.79\% & \textbf{82.29\%} [SD= 0.41] \\
\cmidrule{2-9}
& \multirow{4}{*}{Few-shot} 
    & Foundational & 77.78\% & 78.32\% & 75.99\% & 78.49\% & 78.14\% & 77.74\% [SD= 1.01] \\
    & & Structured & 74.01\% & 74.37\% & 74.55\% & 73.48\% & 73.66\% & 74.01\% [SD= 0.68] \\
    & & Comprehensive & 72.58\% & 71.68\% & 72.22\% & 73.84\% & 73.48\% & 72.76\% [SD= 0.89] \\
    & & Concise & 82.43\% & 82.25\% & 80.82\% & 81.54\% & 81.54\% & \textbf{81.72\%} [SD= 0.64] \\ 
\cmidrule{2-9}
& \multirow{4}{*}{Zero-shot RAG} 
    & Foundational & 79.03\% & 80.29\% & 80.65\% & 78.32\% & 77.06\% &\textbf{79.07\%} [SD= 1.46] \\
    & & Structured & 76.88\% & 75.99\% & 75.27\% & 77.78\% & 78.49\% & 76.88\% [SD= 1.30] \\
    & & Comprehensive & 74.55\% & 74.73\% & 74.01\% & 73.66\% & 75.09\% & 74.41\% [SD= 0.57] \\
    & & Concise & 76.29\% & 75.93\% & 76.32\% & 76.16\% & 75.49\% & 76.03\% [SD= 0.34] \\
\cmidrule{2-9}
& \multirow{4}{*}{Few-shot RAG} 
    & Foundational & 80.87\% & 81.95\% & 82.13\% & 79.93\% & 83.33\% & 81.64\% [SD= 1.30] \\
    & & Structured & 80.32\% & 79.06\% & 80.51\% & 79.93\% & 80.65\% & 80.09\% [SD= 0.63] \\
    & & \cellcolor{highlightblue} Comprehensive & \cellcolor{highlightblue} 82.31\% & \cellcolor{highlightblue} 81.59\% & \cellcolor{highlightblue} 82.31\% & \cellcolor{highlightblue} 81.18\% & \cellcolor{highlightblue} 83.51\% & \cellcolor{highlightblue} \textbf{82.18\%} [SD= 0.88] \\
    & & Concise & 81.18\% & 80.82\% & 80.29\% & 82.36\% & 79.93\% & 80.92\% [SD= 0.93] \\
\midrule
\multirow{6}{*}{\rotatebox{90}{\textbf{\textit{Llama}}}}
& \multirow{2}{*}{Zero-shot}
& Concise & 81.58\% & 82.44\% & 80.98\% & 81.98\% & 81.85\% & \textbf{81.77\%} [SD= 0.53] \\
& & Comprehensive & 77.99\% & 77.52\% & 78.04\% & 77.67\% & 78.06\% & 77.86\% [SD= 0.24] \\
\cmidrule{2-9}
& \multirow{2}{*}{Few Shot}
& \cellcolor{highlightorange} Concise & \cellcolor{highlightorange} 82.99\% & \cellcolor{highlightorange} 83.01\% & \cellcolor{highlightorange} 83.34\% & \cellcolor{highlightorange} 83.20\% & \cellcolor{highlightorange} 81.82\% & \cellcolor{highlightorange} \textbf{82.87\%} [SD= 0.58] \\
& & Comprehensive & 77.84\% & 80.02\% & 79.35\% & 78.68\% & 75.59\% & 78.30\% [SD= 1.71] \\
\cmidrule{2-9}
& \multirow{2}{*}{Zero shot RAG}
& Concise & 81.22\% &  81.02\% & 81.18\% & 81.22\% & 81.64\% & \textbf{81.22\%} [SD= 0.20] \\
& & Comprehensive & 75.31\% & 74.81\% & 74.64\% & 75.43\% & 75.26\% & 75.09\% [SD= 0.34] \\
\cmidrule{2-9}
& \multirow{2}{*}{Few Shot RAG}
& Concise & 81.62\% & 81.23\% & 81.46\% & 81.67\% & 81.21\% & \textbf{81.44\%} [SD= 0.21] \\
& & Comprehensive & 77.32\% & 77.07\% & 77.46\% & 77.11\% & 77.46\% & 79.57\% [SD= 0.37] \\
\midrule
\multirow{6}{*}{\rotatebox{90}{\textbf{\textit{Qwen}}}}
& \multirow{2}{*}{Zero shot}
& Concise & 76.49\% & 76.50\% & 76.49\% & 76.26\% & 76.87\% & \textbf{76.52\%}  [SD= 0.18] \\
& & Comprehensive & 72.07\% & 71.10\% & 71.10\% & 70.82\% & 72.70\% & 71.56\%  [SD= 0.79] \\
\cmidrule{2-9}
& \multirow{2}{*}{Few Shot}
& \cellcolor{highlightblue} Concise & \cellcolor{highlightblue} 84.15\% & \cellcolor{highlightblue} 82.28\% & \cellcolor{highlightblue} 82.29\% &  \cellcolor{highlightblue} 81.96\% & \cellcolor{highlightblue} 82.62\% & \cellcolor{highlightblue} \textbf{82.66\%}  [SD= 0.84] \\
& & Comprehensive & 76.04\% & 76.55\% & 74.41\% & 74.74\% & 77.33\% & 75.81\%  [SD= 1.30] \\
\cmidrule{2-9}
& \multirow{2}{*}{Zero shot RAG}
& Concise & 74.89\% & 74.15\% & 74.15\% & 74.42\% & 74.34\% & \textbf{74.39\%} [SD= 0.30] \\
& & Comprehensive & 70.21\% & 70.70\% & 70.61\% & 70.53\% & 71.04\% & 70.62\%  [SD= 0.29] \\
\cmidrule{2-9}
& \multirow{2}{*}{Few Shot RAG}
& Concise & 82.01\% & 81.29\% & 81.64\% & 81.43\% & 81.81\% & \textbf{81.64\%}  [SD= 0.28] \\
& & Comprehensive & 80.14\% & 79.31\% & 79.70\% & 79.25\% & 79.45\% & 79.57\%  [SD= 0.36] \\  
\hline
\noalign{\vskip 5pt}
\end{tabular}
\caption{F1 Score Results for Sentiment Classification Using Different Prompt Strategies with ChatGPT, Llama, and Qwen.   \\      Note: Bold = best within strategy. {\color{highlightblue}\rule{1.5ex}{1.5ex}} = overall best per model. {\color{highlightorange}\rule{1.5ex}{1.5ex}} = selected for annotation.}
\label{tab:combined_results_sentiment}
\end{table*}

\begin{itemize}
    \item \textbf{Zero-Shot Setting} -- The concise prompt outperformed all other prompts, with an average F1 score of 82.29\%. The foundational, structured, and comprehensive prompt variants were followed with 79.43\% and 77.24\%, 75.44\%, respectively. These results show that presenting essential information in a compact form helps the model to generalize sentiment categories more effectively by reducing cognitive overload.
    \item \textbf{Few-Shot Setting} -- 40\% of the training data were used as examples labeled in the context. The data set was partitioned into five equal subsets; in each run, two subsets formed prompt examples, and the remaining three served as the evaluation set. This rotating scheme ensured that all sentences were evaluated while maintaining a strict separation between training and test. The concise prompt again delivered the highest F1 score at 81.72\%, followed by the foundational (77.74\%), structured (74.01\%), and comprehensive prompt (72.76\%). These results suggest that prompt variants benefit from few-shot examples, overly detailed prompts can dilute the model’s focus on key sentiment cues, whereas concise, focused prompts support better generalization.
    \item \textbf{Zero-Shot RAG Setting} -- The foundational, structured, concise, and comprehensive prompts achieved F1 scores of 79.07\%, 76.88\%, 76.03\%, and 74.41\%, respectively. Overall performance declined in this setting due to retrieval noise, introduces irrelevant information that interfered with prompt clarity and focus.
    \item \textbf{Few-Shot RAG Setting} -- In the few-shot RAG setting, comprehensive, concise, foundational and structured prompts achieved F1 scores of 82.18\%, 80.92\%, 81.64\%, and 80.09\%, respectively. This was the only setting comprehensive prompt to show a clear advantage, while the concise prompt remained highly competitive and consistently effective across all configurations.
\end{itemize}

To further examine the influence of prompt design on model performance, we conducted a cross model comparison using the two most competitive strategies, concise and comprehensive prompt. The results highlight how different prompt styles interact with specific language models and evaluation setups.

\begin{itemize}
    \item \textbf{Performance on Llama} -- In the zero shot condition, concise and comprehensive prompts achieved F1 scores of 81.77\% and 77.86\%, respectively. The zero shot RAG setting showed similar trends, with concise (81.22\%) and comprehensive prompt (75.09\%). In the Few-shot setting, concise (82.87\%) and comprehensive prompt (78.30\%). In the few-shot RAG condition, concise (81.44\%), outperforming the comprehensive prompt (79.57\%). Across all conditions, the concise prompt consistently outperformed the comprehensive prompt, with only modest performance differences, reinforcing that semantically focused prompts are more robust to retrieval noise.
    \item \textbf{Performance on Qwen} -- In the zero shot setting, concise achieved 76.52\%, while the comprehensive prompt achieved 71.56\%. The zero-shot RAG configuration resulted in similar behavior: Concise yielded 74.39\% and 70.62\% with a comprehensive prompt. Concise achieved 82.66\%, substantially higher than comprehensive prompt at 75.81\%. In the Few-shot RAG setting, concise also outperformed the comprehensive prompt (81.64\% and 75.96\%). These findings indicate that Qwen, consistently benefits from concise, high-density semantic prompts, and that longer prompts with added background or definitions, as used in comprehensive prompt.
\end{itemize}

\subsubsection{Comparison between Different Prompt Strategies for Sentiment Analysis}

\begin{itemize}
    \item \textbf{Prompt Efficiency:} We define prompt efficiency as the interplay between performance, token and computational cost (e.g., number of examples, retrieval overhead), and human involvement in crafting prompts. Prompt variants varies in complexity: foundational prompts are simple, structured prompts add brief human authored definitions, comprehensive prompts include extended context, and concise prompts are derived by summarizing comprehensive prompts using an LLM with expert review increasing the human effort, retaining essential information. The zero shot requires the least human and token overhead, relying on model achieving moderate performance (ChatGPT 82.29\%, Llama 81.77\%, Qwen 76.52\%; SD 0.18–0.53). Its simplicity makes it efficient but generally outperformed by guided strategies. Few shot prompting improves performance by supplying curated examples at the cost of higher token usage and manual effort. Few-shot achieves the highest scores (Llama 82.87\%, Qwen 82.66\%, ChatGPT 81.72\%) and improves model understanding for sentiment task with manageable overhead. Retrieval Augmented Generation (RAG) introduces external knowledge by adding retrieved documents to the prompt. Zero shot RAG yields mixed results: Llama 81.22\% (SD 0.20) performs stably, while ChatGPT 79.07\% (SD 1.46) shows higher variance, reflecting sensitivity to retrieval quality. Few-shot RAG, which combines examples and retrieval, achieves strong but only marginally higher performance than standard few shot (ChatGPT 81.29\%, SD 0.93) with greater computational and retrieval costs due to longer prompts and external queries. In this context, few shot prompting provides the best balance of accuracy and resource cost, zero shot remains the lightest weight option with moderate accuracy, and few-shot RAG offers minimal additional gains at significantly higher computational expense. Optimal strategy selection should therefore weigh the performance benefits against token usage, retrieval overhead, and human effort in crafting prompts.
    \item \textbf{Generalizability:} We define generalizability as the ability of a prompt strategy to maintain consistent performance in different LLM architectures. The few shot concise strategy demonstrates strong generalizability, achieving top performance across all models ChatGPT 81.72\%, Llama 82.87\%, and Qwen 82.66\%, with low variance and closely aligned F1 scores. In contrast, zero shot strategies show greater fluctuation: the concise zero shot prompt reaches 82.29\% in ChatGPT and 81.77\% in Llama but drops to 76.52\% in Qwen, indicating weaker generalization. RAG-based strategies also vary; zero shot RAG (concise) performs well in Llama 81.22\% but less consistently in ChatGPT 76.29\% and Qwen 76.52\%, reflecting sensitivity to retrieved context alignment. Overall, few shot concise emerges as the most generalizable approach, offering high and stable accuracy across models, whereas retrieval based and purely zero-shot formats are more sensitive to model architecture.
    \item \textbf{Retrieval \& Noise Sensitivity:} We define retrieval and noise sensitivity as a strategy’s susceptibility to performance fluctuations due to irrelevant or unstable retrieved content. Across models, RAG based strategies exhibit varying robustness. The zero shot RAG in ChatGPT shows a high standard deviation (SD = 1.46) and moderate F1 score (79.07\%), indicating sensitivity to noisy retrieval. In contrast, llama's zero shot RAG is highly stable (SD = 0.20, F1 score = 81.22\%), suggesting better alignment with retrieved inputs. Few shot RAG strategies also show mixed behavior: ChatGPT (SD = 0.93) and Qwen (SD = 0.26) perform slightly worse than their standard few shot counterparts despite added context, implying that retrieval sometimes introduces noise rather than clarity. Overall, retrieval augmented strategies are more sensitive to input quality, and their benefit is model dependent highlighting the need to carefully manage retrieved content to avoid performance degradation.
    \item \textbf{Prompt Stability:} is measured as the standard deviation (SD) across five test runs, reflects how consistently a strategy performs under repeated execution. Few shot strategies exhibit the most stable behavior overall particularly with llama (few shot RAG concise: SD = 0.21) and Qwen (few shot RAG concise: SD = 0.25). In contrast, zero shot RAG shows higher variance, especially in ChatGPT (SD = 1.46), suggesting greater sensitivity to fluctuating retrieval inputs. The zero shot and standard Few-shot remain relatively stable across models, with SDs typically below 0.70. Overall, few shot RAG (concise) consistently yields the lowest variance, indicating strong stability when both examples and retrieved context are well aligned with the model.
    
\end{itemize}

\subsubsection{Comparison between Different LLMs for Sentiment Analysis}

\begin{itemize}
    \item \textbf{Accuracy:} Sentiment analysis was performed on ChatGPT, Llama, and Qwen, all demonstrating strong and closely aligned performance, Llama achieved the highest accuracy (82.87\%, few shot concise), followed by Qwen (82.66\%) and ChatGPT (81.72\%). The close performance across models shows, sentiment task is simple and that current LLMs can achieve high accuracy reliably with minimal instruction.
    \item \textbf{Stability:} Llama shows the highest stability, showing the lowest standard deviations and reduced sensitivity to prompt variability, which makes it well suited for stable sentiment analysis. ChatGPT exhibits greater fluctuation, reaching its peak at SD=1.46 under zero shot RAG foundational, indicating greater prompt sensitivity. Despite similar accuracy across models, Llama combines high performance with low variance, underscoring its robustness for minimally instructed sentiment classification.

    \item \textbf{Prompt Length:} Prompt length significantly affects the stability of the model. Across all models, shorter prompts (concise) consistently yield lower variance than longer ones (comprehensive). Llama shows exceptional stability with concise prompts(SD=0.20), zero shot RAG (SD=0.21), and few shot RAG compared to SD=1.71 for its comprehensive variant. The same pattern is observed in ChatGPT (concise: SD = 0.41–0.64, comprehensive: up to SD = 0.88) and Qwen (concise: SD = 0.18–0.84, comprehensive: SD = 1.46). These results suggest that longer prompts introduce greater instability, due to increased token-level sensitivity and over-specification, while shorter prompts improve consistency that are well optimized for minimal input.
    
    \item \textbf{Transferability:} To ensure consistency and cross model comparability, we adopted a prompt transferability approach, where all prompts were initially developed using ChatGPT and then applied without modification to Llama and Qwen. This setup enables an effective evaluation of how well prompts generalize across models. The results show only minor variations in performance, with all models achieving closely aligned F1 scores. The well-structured prompts are highly transferable, enabling reproducible evaluations across LLMs. For historical contexts such as Japanese American incarceration narratives, transferability provides a robust framework for comparative sentiment analysis, minimizing bias from prompt design and enhancing scientific reproducibility.
\end{itemize}

These findings highlight that concise, well structured prompts allow optimizing performance across models and also enhance stability and generalizability, offering a reliable foundation for large-scale sentiment analysis in historically sensitive corpora.

\subsection{Semantic Classification}

We evaluated the effectiveness of five systematically developed prompt variants - Foundational, Structured, Comprehensive, Refined, and Concise across four experimental conditions: Zero-shot, Zero-shot RAG, Few-shot, and Few-shot RAG. These prompt variants were designed to incrementally enrich the semantic scaffolding provided to the model, the results are presented in Table~\ref{tab:combined_results_semantic}.  Each variant was tested five times across the entire data set to maintain the consistency and reliability of the results.

\begin{table*}[htbp]
\centering
\small
\begin{tabular}{cllcccccc}
\toprule
\multirow{2}{*}{\textbf{Model}} & 
\multicolumn{1}{c}{\multirow{2}{*}{\textbf{Strategy}}} & 
\multicolumn{1}{c}{\multirow{2}{*}{\textbf{Variant}}} & 
\multicolumn{5}{c}{\textbf{Stability Test}} & 
\multirow{1}{*}{\textbf{Average /}} \\
\cline{4-8}
 & & & \textbf{Test 1} & \textbf{Test 2} & \textbf{Test 3} & \textbf{Test 4} & \textbf{Test 5} & \textbf{Standard Deviation [SD]}\\
\hline
\multirow{20}{*}{\rotatebox{90}{\textbf{\textit{ChatGPT}}}}
& \multirow{5}{*}{Zero-shot}
    & Foundational & 53.41\% & 52.33\% & 52.69\% & 50.54\% & 53.05\% & 52.40\% [SD= 0.99] \\
&& Structured & 68.10\% & 66.67\% & 67.20\% & 68.46\% & 69.00\% & 67.89\% [SD= 0.84] \\
&& Comprehensive & 80.47\% & 79.93\% & 79.57\% & 80.11\% & 81.18\% & 80.25\% [SD= 0.54] \\
&& Refined    & 82.18\% & 83.87\% & 83.41\% & 83.36\% & 83.93\% & \textbf{83.35\%} [SD= 0.62] \\
&& Concise    & 82.15\% & 82.15\% & 82.87\% & 83.58\% & 83.22\% & 82.79\% [SD= 0.57] \\

\cmidrule{2-9}
& \multirow{5}{*}{Few-shot}
    & Foundational & 83.64\% & 74.55\% & 78.18\% & 76.36\% & 76.36\% & 77.82\% [SD= 3.12] \\
&& Structured & 78.18\% & 83.64\% & 80.00\% & 74.55\% & 83.64\% & 80.00\% [SD= 3.45] \\
&& Comprehensive & 85.45\% & 80.00\% & 83.64\% & 78.18\% & 80.00\% & 81.45\% [SD= 2.67] \\
&& \cellcolor{highlightblue} Refined    & \cellcolor{highlightblue} 89.43\% & \cellcolor{highlightblue} 87.28\% & \cellcolor{highlightblue} 87.99\% & \cellcolor{highlightblue} 89.07\% & \cellcolor{highlightblue} 89.78\% & \cellcolor{highlightblue} \textbf{88.71\%} [SD= 0.93] \\
&& Concise    & 83.33\% & 84.05\% & 83.33\% & 83.15\% & 83.51\% & 83.47\% [SD= 0.30] \\

\cmidrule{2-9}
& \multirow{5}{*}{Zero-shot RAG}
    & Foundational & 13.08\% & 12.54\% & 13.63\% & 13.26\% & 12.90\% & 13.08\% [SD= 0.36] \\
&& Structured & 44.27\% & 43.01\% & 43.91\% & 44.44\% & 45.34\% & 44.19\% [SD= 0.75] \\
&& Comprehensive & 37.63\% & 36.20\% & 37.46\% & 39.07\% & 39.61\% & 37.99\% [SD= 1.21] \\
&& Refined    & 81.40\% & 82.78\% & 81.40\% & 82.86\% & 82.22\% & \textbf{82.13\%} [SD= 0.63] \\
&& Concise    & 79.57\% & 81.90\% & 81.00\% & 78.14\% & 77.78\% & 79.68\% [SD= 1.59] \\

\cmidrule{2-9}
& \multirow{5}{*}{Few-shot RAG}
    & Foundational & 38.17\% & 45.34\% & 43.37\% & 41.40\% & 38.35\% & 41.33\% [SD= 2.79] \\
&& Structured & 62.19\% & 63.80\% & 61.29\% & 60.57\% & 63.08\% & 62.19\% [SD= 1.16] \\
&& Comprehensive & 79.57\% & 81.90\% & 81.00\% & 78.14\% & 77.78\% & 79.68\% [SD= 1.59] \\
&& Refined    & 87.63\% & 87.63\% & 86.91\% & 85.84\% & 88.53\% & \textbf{87.31\%} [SD= 0.89] \\
&& Concise    & 82.79\% & 83.15\% & 83.51\% & 84.05\% & 81.90\% & 83.08\% [SD= 0.72] \\
\midrule

\multirow{6}{*}{\rotatebox{90}{\textbf{\textit{Llama}}}}
& \multirow{2}{*}{Zero-shot}
& Concise & 79.91\% & 79.01\% & 79.65\% & 80.29\% & 79.74\% & 79.72\% [SD= 0.41] \\
& & Refined & 83.29\% & 82.92\% & 83.48\% & 83.15\% & 82.62\% & \textbf{83.09\%} [SD= 0.29] \\
\cmidrule{2-9}
& \multirow{2}{*}{Few Shot}
&  Concise & 82.62\% & 82.14\% & 85.61\% & 85.03\% & 85.46\% & 84.17\% [SD= 1.48] \\
& & \cellcolor{highlightorange} Refined & \cellcolor{highlightorange} 84.97\% & \cellcolor{highlightorange} 85.19\% & \cellcolor{highlightorange} 84.96\% & \cellcolor{highlightorange} 84.82\% & \cellcolor{highlightorange} 85.00\% & \cellcolor{highlightorange} \textbf{84.99\%} [SD= 0.11] \\
\cmidrule{2-9}
&  \multirow{2}{*}{Zero shot RAG} 
& Concise & 76.94\% & 76.38\% & 76.96\% & 77.02\% & 77.26\% & 76.91\% [SD= 0.41] \\
& & Refined & 82.15\% & 82.72\% & 82.72\% & 83.37\% & 82.49\% & \textbf{82.69\%} [SD= 0.29] \\
\cmidrule{2-9}
& \multirow{2}{*}{Few Shot RAG}
&  Concise & 81.56\% & 80.85\% & 81.16\% & 81.42\% & 80.78\% & 81.15\% [SD= 0.30] \\
& & Refined & 83.90\% & 84.15\% & 84.35\% & 83.92\% & 84.13\% & \textbf{84.09\%} [SD= 0.16] \\
\midrule
\multirow{6}{*}{\rotatebox{90}{\textbf{\textit{Qwen}}}}
& \multirow{2}{*}{Zero shot}
& Concise & 83.00\% & 82.43\% & 82.01\% & 82.05\% & 83.00\% & \textbf{82.49\%} [SD= 0.43] \\
& & Refined & 81.47\% & 81.44\% & 81.79\% & 81.81\% & 81.41\% & 81.58\% [SD= 0.17] \\
\cmidrule{2-9}
& \multirow{2}{*}{Few Shot}
& Concise & 82.78\% & 81.85\% & 82.11\% & 81.48\% & 82.01\% & 82.05\% [SD= 0.42] \\
& & Refined & 82.80\% & 83.10\% & 83.05\% & 83.36\% & 83.28\% & \textbf{83.12\%} [SD= 0.19] \\
\cmidrule{2-9}
& \multirow{2}{*}{Zero shot RAG}
& Concise & 83.22\% & 82.98\% & 82.98\% & 83.45\% & 83.77\% & 83.28\% [SD= 0.30] \\
& & \cellcolor{highlightblue}Refined & \cellcolor{highlightblue} 83.02\% & \cellcolor{highlightblue} 83.45\% & \cellcolor{highlightblue} 83.77\% & \cellcolor{highlightblue} 84.57\% & \cellcolor{highlightblue} 83.77\% & \cellcolor{highlightblue} \textbf{83.72\%} [SD= 0.50] \\
\cmidrule{2-9}
& \multirow{2}{*}{Few Shot RAG}
& Concise & 81.15\% & 83.50\% & 83.47\% & 82.07\% & 81.72\% & \textbf{82.38\%} [SD= 0.94] \\
& & Refined & 82.32\% & 82.44\% & 81.60\% & 82.03\% & 81.15\% & 81.91\% [SD= 0.47] \\ 
\bottomrule
\noalign{\vskip 5pt}
\end{tabular}
\caption{F1 Score Results for Semantic Classification Using Different Prompt Strategies with ChatGPT, Llama, and Qwen       \\      \textbf{Note:} Bold = best within strategy. {\color{highlightblue}\rule{1.5ex}{1.5ex}} = overall best per model. {\color{highlightorange}\rule{1.5ex}{1.5ex}} = selected for annotation.}
\label{tab:combined_results_semantic}
\vspace{-2.5em}
\end{table*}

\subsubsection{Evaluation of LLM Performance for Semantic Classification}\hfill\\
We begin with ChatGPT, as it represents a widely adopted foundation model whose behavior offers a reference point for evaluating the impact of prompt design before extending the analysis to other open source models.

\begin{itemize}
    \item \textbf{Zero shot setting} -- Model performance increased as prompts became more structured: the foundational prompt achieved an F1 score of 52.40\%, the structured prompt 67.89\%, and the comprehensive prompt 80.25\%. The refined prompt reached 83.35\%, a +30.95\% improvement over the foundational prompt, while the concise version achieved 82.79\%, showing that shorter, focused prompts can perform as well as longer ones.
    \item \textbf{Few shot Setting} -- In the few-shot setting, the foundational, structured, and comprehensive prompts achieved 77.82\%, 80.00\%, and 86.15\%, respectively. The refined prompt reached 88.71\%, while the concise prompt achieved 83.47\%, providing a +5.24\% gain over the foundational prompt with reduced length.
    \item \textbf{Zero shot RAG Setting} -- The foundational prompt performed poorly with an F1 score 13.08\%, with structured and comprehensive prompts reaching 44.19\% and 37.99\%, respectively. The refined prompt achieved 82.13\%, a +69.05\% improvement over the foundational prompt, while the concise prompt scored 79.68\%, showing that well focused prompts can reduce errors caused by RAG. 
    \item \textbf{Few shot RAG Setting} -- In the context rich few shot RAG setting, the foundational prompt F1 score was 41.33\%, improving to 62.19\% with the structured prompt and 79.68\% with the comprehensive prompt. The refined prompt achieved the highest F1 score of 87.31\%, followed by the concise prompt at 83.08\%, showing that focused prompts yield the most reliable performance.
\end{itemize}

To evaluate the generalizability and robustness of our prompt engineering strategies, we tested the two highest-performing prompt variants Refined and Concise on two additional open-source LLMs: Llama and Qwen. These experiments were designed to assess whether the performance trends observed with ChatGPT extend to different model architectures and pretraining paradigms.

\begin{itemize}
    \item \textbf{Performance on Llama} -- The results show that the refined prompt consistently outperformed the concise prompt in all variants of the prompt. The zero shot setting, refined achieved an avg F1 score of 83.09\%, compared to 79.72\% with Concise. In the zero shot RAG setting, refined with 82.69\%, concise achieved 76.91\%, suggesting that detailed and semantically grounded prompts help mitigate the noise introduced by retrieval. In the few shot, refined achieved 84.99\%, slightly higher than concise at 84.17\%, indicating that both prompt variants are competitive under this setup. In the few shot RAG setting, refined maintained its lead with 84.09\%, while concise with 81.15\%. These results suggest that the llama benefits from structured and detailed prompts for semantic classification, allowing the model to effectively resolve ambiguity in both standard and RAG settings.
    \item \textbf{Performance on Qwen} -- In the zero shot setting, concise achieved the highest performance at 82.49\%, surpassing refined 81. 58\%. In the zero shot RAG, both prompts performed close, with concise 83.28\% and refined at 83.72\%, reflecting the model’s resilience to contextual noise. In the few shot setting, the refined slightly outperformed the concise (82.68\% vs. 82.05\%), although the difference was marginal. However, in the few shot RAG setting, concise achieved 82.38\% and refined at 81.58\%. These results show Qwen is better able to utilize compact and semantically dense prompts, due to characteristics of its architecture, tokenizer behavior, that make it more effective with concise contextual input.
\end{itemize}


\subsubsection{Comparison between Different Prompt Strategies for Semantic Classification}

\begin{itemize}
    \item \textbf{Prompt Efficiency:} Each prompt variant reflects a trade-off between performance and efficiency. Refined prompts achieve the highest accuracy (ChatGPT few-shot refined 88.71\%, SD=0.93) but incur higher computational cost. Concise prompts, shortened by LLMs and verified by humans, deliver strong performance with lower token usage (Llama few-shot concise 84.17\%, SD=0.15 vs. refined 84.99\%, SD=0.11), making them more cost effective but wth further refinement which require human intervention.

    Zero shot prompting is highly sensitive to prompt quality, benefiting from clear structure and LLM-assisted refinement. Few shot prompting is more stable but gains from optimized designs. In zero-shot RAG, retrieval boosts performance but diminishes the relative benefit of prompt refinement, while few shot RAG consistently delivers the best results, where refined and concise prompts excel through the combined effect of optimized instructions and contextual augmentation. Prompt efficiency depends not only on accuracy but also on achieving it with minimal tokens, computation, and human effort. Our results underscore the value of using LLMs not only as inference engines but also as collaborators in prompt creation and compression. The refined and concise LLM prompts strike an optimal balance, achieving high semantic classification performance with minimal resource consumption.
    \item \textbf{Generalizability:} The ability to maintain stable and high performing results across the models is important. Rather than evaluating the models themselves, our analysis centers on whether a specific strategy and variant performs consistently across ChatGPT, Llama, and Qwen. For instance, the few shot Refined strategy yields comparably strong results across all three models chatGPT (88.71\%), Llama (84.99\%), and Qwen (84.93\%) with low standard deviation in each case, indicating high stability. In contrast, other strategies, such as zero-shot, show greater variability in performance across models, suggesting that they are more sensitive to architectural differences. These findings emphasize that prompting strategies vary in their generalizability, and those that incorporate model support and LLM optimized prompt refinement tend to be more robust.  
    \item \textbf{Retrieval \& Noise Sensitivity:} In the context of semantic classification on historical context, our results show that RAG based strategies are particularly sensitive to retrieval noise, especially under foundational and structured prompts in the few shot RAG setting. These prompts often fail to constrain the model’s reliance on irrelevant related retrieved content, leading to unstable performance. By contrast, refined and concise prompts consistently improve stability (e.g., ChatGPT Refined Few shot RAG: 87.31\%), showing that well designed prompts reduce semantic drift and enhance retrieval robustness. This highlights the importance of evaluating noise sensitivity, particularly when working with complex, historically nuanced datasets. Careful prompt engineering is essential to ensure consistent and reliable performance in RAG-based semantic classification. 
    \item \textbf{Prompt Stability:} Prompt stability measured by standard deviation across repeated runs reveals important insights into the reliability of semantic classification. Few shot prompting generally yields higher standard deviation than zero shot, particularly for foundational and structured variants (e.g., ChatGPT few shot Structured: SD = 3.45, vs. Refined: SD = 0.93). This suggests that longer or under specified prompts are more susceptible to instability, due to ambiguous and inconsistent internal reasoning by the model across runs. In contrast, refined and concise variants consistently show low standard deviation, indicating strong prompt stability (e.g., Llama Refined Few-shot: SD = 0.11, Qwen Refined Zero-shot: SD = 0.19). This underscores the importance of compact, well-scoped prompt design to minimize stochastic variation and improve reproducibility. To reduce prompt instability, especially when working with historical or nuanced text, practitioners should avoid verbose structured prompts and instead focus on clear, domain aligned, and task specific formulations. Incorporating prompt regularization and repeat run evaluation can further ensure consistent outcomes.

\end{itemize}

\subsubsection{Comparison between Different LLMs}

\begin{itemize}
    \item \textbf{Accuracy:} Across all prompting strategies, ChatGPT achieves the highest semantic classification accuracy, particularly under the refined few-shot variant (88.71\%). Llama and Qwen perform competitively (e.g., Llama refined few shot 84.99\%, Qwen refined few-shot RAG 83.72\%) but remain below ChatGPT’s peak performance. This reflects the influence of architecture, scale, and instruction tuning in capturing nuanced semantic boundaries. Model complexity and design are critical considerations when selecting LLMs for domain-specific tasks. While ChatGPT offers superior results, it may be less practical in resource-constrained settings. Llama and Qwen provide strong performance with lower variance and are often preferable for efficiency-focused applications. Effective LLM selection for semantic tasks should balance accuracy, model size, and inference cost.
    \item \textbf{Stability:} When evaluating the stability of the model measured by the standard deviation between repeated runs, the llama emerges as the most stable large language model overall. Its refined variant with few shots achieves an accuracy of 84.99\% with just SD=0.11, indicating highly consistent performance. Similarly, Llama's other top performing variants exhibit low variance (e.g., zero shot refined: SD=0.29), reinforcing its robustness. In comparison, ChatGPT, while delivering the highest overall accuracy (e.g., refined few shot: 88.71\%, SD=0.93), shows greater variability across test runs, particularly in foundational and structured prompting styles (SD > 3.0). Qwen demonstrates a balance between performance and stability, with multiple configurations showing SD<0.5 (e.g., Refined Few-shot: SD = 0.30). These results suggest that, while ChatGPT is strongest in peak performance, Llama offers more consistent output, making it suitable for tasks where predictability and reproducibility are critical. Model selection should therefore consider not only mean performance but also variance across runs, especially for deployment in sensitive or production environments.
    \item \textbf{Prompt Length:} Prompt length affects both accuracy and stability in LLM based semantic classification. Structured and comprehensive prompts add context, show moderate performance gains but higher variability (e.g., ChatGPT few-shot structured: 80.00\%, SD=3.45). Refined prompts, which are longer and carefully optimized, achieve the highest accuracy with low variance (e.g., ChatGPT few-shot refined: 88.71\%, SD=0.93; Llama few shot refined: 84.99\%, SD=0.11), showing that clarity and task specificity are more important than length alone. Concise prompts, despite being shorter, also maintain strong performance with low variability, demonstrating that well-scoped, high-density instructions can match or approach the effectiveness of more detailed prompts.
    
    These findings show that effective prompt design requires balancing length, clarity, and token usage. Longer prompts increase computational cost and can cause semantic drift if poorly structured. Prompt length alone does not determine performance; semantic precision and clear instructions are essential for achieving high accuracy and stability, particularly in token limited or deployment sensitive settings.
    \item \textbf{Transferability:} While prompts were initially designed using ChatGPT, their transferability to other models namely Llama and Qwen in semantic classification shows notable limitations. Although ChatGPT achieves the highest accuracy (e.g., 88.71\% with refined few shot), performance drops when the same prompts are applied to Llama (84.99\%) and Qwen (83.72\%), highlighting a loss in cross model generalizability. This gap is driven by three factors: (1) ChatGPT-specific instruction tuning, which aligns well with task formatted prompts; (2) architectural differences in how models interpret complex or multi-intent instructions; and (3) stability variance. Compared to sentiment analysis, which exhibited minimal or negligible performance degradation across models, semantic classification is more sensitive to prompt transfer, due to its higher conceptual complexity and dependence on fine-grained category boundaries. These findings suggest that while prompt based methods are portable, semantic transferability is less reliable than for more bounded tasks like sentiment detection emphasizing the need for model specific prompt adaptation when deploying prompts beyond their source model.
\end{itemize}



\subsection{Exploring Sentiment and Semantic Structure in Oral History with LLMs}




\subsubsection{LLM for Sentiment Analysis in Oral History Archive}\hfill\\
Clear trends emerge across prompt variants and models for sentiment analysis in oral history analysis. For ChatGPT, the foundational prompt yielded strong performance, showing the inherent understanding of ChatGPT's sentiment. The structured prompt to add more examples or context did not always improve the performance of a few shots. The concise prompt, which uses fewer tokens, achieved the highest scores for both zero shot and few shot, showing that compact prompts can outperform longer ones for sentiment analysis. Llama and Qwen, the trend shifted. Few shot strategy with concise prompting produced the highest scores in the prompt variants. Unlike ChatGPT, these models benefited more from explicit examples, highlighting their dependence on well structured, informative input. The effectiveness of concise prompts in all models reinforces the value of prompt efficiency, but the different strategies between ChatGPT and the other models suggest that optimal prompts must be tailored to the capabilities of each model. This suggests that minimal guidance can yield meaningful results, and fine-tuning offers further gains. Initial prompt variants, particularly under zero-shot and RAG based strategies, exhibit greater variability. This suggests that early stage prompt formulations were more sensitive to variations across runs, due to less stable model behavior or inconsistent retrieval relevance. In contrast, refined prompts, especially concise variants, consistently have lower standard deviations across all models, with Llama and Qwen often falling low deviation. These findings underscore that careful, efficient prompt engineering reduces output variance and that standard deviation complements average F1 in evaluating prompt effectiveness.

Oral history sentiment analysis poses unique interpretive challenges. Oral histories are emotionally layered and culturally situated, requiring prompts that capture this richness. The success of concise prompts across models highlights the importance of designing instructions that are sensitive but minimally intrusive. High performing models still risk errors when emotional tone is implicit or culturally nuanced, as in Japanese-American incarceration narratives. Therefore, prompt design must balance clarity with contextual subtlety, avoiding overly simplistic definitions that may flatten complex human experiences. Model selection should also consider historical and ethical alignment. To prevent distortions in large scale annotation, researchers must iteratively test prompts, verify model behavior on edge cases, and consult domain experts. Ultimately, ensuring the integrity of oral history sentiment analysis requires a careful synthesis of computational rigor and humanistic judgment.





\subsubsection{LLM for Semantic Analysis in Oral History Archive}\hfill\\
Semantic classification requires domain understanding and is highly sensitive to prompt structure. This task introduced a refined prompt to better capture semantic nuance. With ChatGPT, we observed a performance range, from 13.08\% to 88.71\% F1 score, reflecting a 75 point differential that underscores how critical prompt design is for complex classification tasks. Although concise prompts proved effective in sentiment analysis, they did not generalize as well in the semantic setting. Our findings highlight that prompt design is not one-size-fits-all, particularly across tasks such as sentiment and semantic classification. The refined prompt consistently achieved the highest performance, especially with the llama, which also showed remarkable stability, with precision varying slightly between 82.69\% and 84.9\%. These findings reinforce that semantic classification benefits from detailed, context aware prompts and that llama offers strong and consistent performance, making it a reliable choice. Semantic analysis showed higher standard deviation than sentiment, reflecting greater complexity and variability (Table~\ref{tab:combined_results_semantic}). The refinement and clarification of the results significantly reduced this variance. Among all configurations, the few-shot strategy with refined prompt yielded the lowest standard deviation and the best F1 score. Based on both the F1 score and the consistency, this setup was selected for the final annotation of the data.

Across both sentiment and semantic classification, we observed that concise prompts often performed competitively with refined prompts, achieving strong results with lower token usage, reducing inference cost. The nature of the task matters significantly. Sentiment analysis, being more general and well represented in pre-trained LLMs, benefited from concise prompts. In contrast, semantic classification, requires more domain specific, required richer, structured prompts. This distinction underscores that a single prompt formulation cannot generalize across task types and that prompt design must be aligned with the semantic depth and contextual demands of the task. Moreover, our experiments show that few-shot and RAG settings can introduce variability, particularly when the prompt structure lacks clarity. While RAG offers flexibility, it can also amplify noise if it is not paired with carefully crafted prompts. We highlight the value of human and LLM collaboration in rapid development. By combining manual expertise with LLM-generated refinements, we were able to create well-structured and semantically rich prompts, leading to improved model performance. This human-in-the-loop approach to prompt engineering is key to achieving robust and adaptable LLM pipelines in diverse NLP tasks. This suggests that LLMs can not only perform classification tasks, but also assist in generating efficient prompt formulations, offering a scalable solution for large-scale deployments.

\section{Applications of LLMs in Understanding Japanese American Incarceration Oral History}


\subsection{Automated Collection Annotation}

Based on our prompt evaluation results, we selected the concise prompt with few shot strategy for sentiment analysis using llama and the refined prompt with few shot strategy for semantic classification using llama. These configurations were used to automatically annotate the full corpus of 92,191 sentences using python based API calls.

\begin{table}[htbp]
\centering
\vspace{-0.5em}
\begin{tabular}{lrrrrrrr}
\toprule
 & \textbf{Class 0} & \textbf{Class 1} & \textbf{Class 2} & \textbf{Class 3} & \textbf{Class 4} & \textbf{Class 5} & \textbf{Total} \\
\midrule
\textbf{Positive}   & 2,267   & 5,262   & 3,325   & 1,130   & 1,034   & 979     & \textbf{13,997} \\
\textbf{Neutral}    & 18,336  & 13,195  & 13,379  & 5,778   & 2,723   & 1,196   & \textbf{54,607} \\
\textbf{Negative}   & 1,445   & 4,025   & 13,339  & 2,273   & 1,339   & 1,166   & \textbf{23,587} \\
\midrule
\textbf{Total}      & \textbf{22,048} & \textbf{22,482} & \textbf{30,043} & \textbf{9,181} & \textbf{5,096} & \textbf{3,341} & \textbf{92,191} \\
\bottomrule
\noalign{\vskip 5pt}
\end{tabular}
\caption{Distribution of Sentences Across Sentiment Categories and Semantic Classes}
\label{tab:sentence_distribution}
\vspace{-0.5cm}
\end{table}

To understand Japanese-American incarceration narratives, we applied a large-scale category annotation process using a refined few-shot prompting strategy with the Llama model. The model was selected based on its superior performance in multiple prompts and models. The final distribution of the classifications across 92,191 filtered sentences is presented in Table~\ref{tab:sentence_distribution} of sentiment across semantic categories in the JAIOH collection. Class 0 (Biographical Information) is overwhelmingly neutral (18,308 sentences), as expected for content focused on factual background and family history, with minimal positive (2,271) or negative (1,469) sentiment. In Class 1 (Life Before Removal), neutral sentiment (13,121) remains dominant, but the presence of more negative (4,025) than positive (3,336) sentences reflects the discrimination and adversity faced before incarceration. Class 2 (Life During Incarceration) contains the highest number of overall negative sentences (13,233), with fewer positive sentences (3,432) and a comparable number of neutral sentences (13,376), underscoring the emotional hardship and difficult conditions associated with internment. Class 3 (Military Service) exhibits relatively more positive sentiment (1,130) than negative (1,281), indicating pride in military contributions despite historical injustices. Class 4 (Returning After WWII) and 5 (Movements for Peace and Justice) show a more balanced sentiment distribution. Overall, the results suggest that the emotional tone varies meaningfully between narrative themes, with personal hardship and injustice often associated with more negative sentiment, while descriptive or aspirational content tends to be more neutral.




 The analysis reveals that Class 2 (Life During Incarceration) is the most prevalent category, comprising 30,041 sentences. This reflects the emphasis of the narrators on daily life within incarceration camps, including living conditions, emotional experiences, and social dynamics. Class 1 (Life Before Removal) and Class 0 (Biographical Information) follow closely, with 22,482 sentences (22.4\%) and 22,048 sentences (22\%), respectively. These three categories together account for approximately 74\% of all labeled content, suggesting that personal identity, preincarceration life, and confinement represent the core narrative structure in the interviews. However, post-incarceration themes are less frequently discussed. Class 3 (Military Service) appears in 9,181 sentences (9.2\%), Class 4 (Returning After WWII) in 5,096 sentences (5.1\%), and Class 5 (Peace and Justice Movements) in 3,341 sentences (3.3\%). The significant drop from Class 2 to Classes 4 and 5 highlights the dominant narrative weight placed on the camp experience, with fewer narrators elaborating on reintegration, activism, or redress efforts. This distribution reflects the emotional significance of internment, and also the structural emphasis shaped by both the interviewer's focus and collective memory.

\subsection{Entity Extraction and Analysis}

We used the DSLIM/BERT-base-NER \footnote{\url{https://huggingface.co/dslim/bert-base-NER}}  model for its robust performance in English language texts \cite{chadda2024ai} that involve culturally specific entities, making it effective for extraction, and institutions from Japanese-American incarceration narratives. To enhance coverage of domain-specific references, we developed a hybrid entity extraction pipeline that combined this transformer-based model with curated pattern matching rules tailored to incarceration-related entities. This balances broad linguistic coverage with high recall for culturally significant terms. Entities were extracted at the sentence level from 92,191 oral history sentences, each linked to topical categories and sentiment classes. These results were aggregated for frequency analysis to support corpus enrichment and sociocultural interpretation.

\begin{figure}[htbp]
    \centering
    \includegraphics[width=1.0\textwidth]{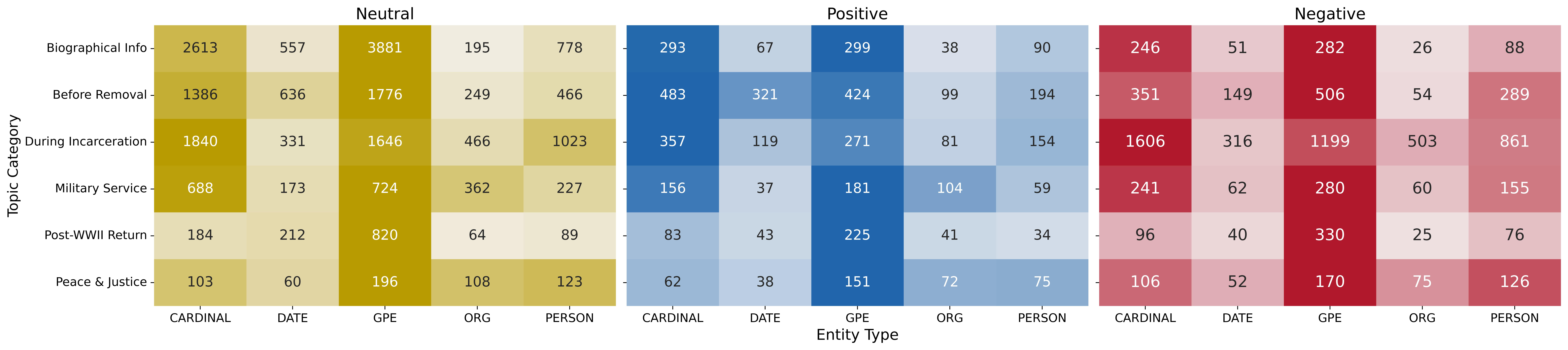}
    \vspace{-2em}
    \Description{Heatmap visualizing the frequency of named entities by sentiment polarity and topic category.}
    \caption{Entity frequency heatmaps by sentiment and topic category from the complete annotated dataset.}
    \label{fig:Entity_Heatmap}
\end{figure}

This analysis provides a class wise view of how the narrators referenced people, places, institutions, and time in their stories. As shown in the entity frequency heatmaps (Figure~\ref{fig:Entity_Heatmap}), PERSON and GPE are especially prominent in neutral and positive narratives, reflecting the role of individual memory and geographic grounding, while the entities ORG, CARDINAL and DATE appear more selectively in institutional and event-based contexts. Each topic class reveals a distinct set of linguistic cues and cultural references shaped by historical experience and personal recollection. The distribution of named entities not only reinforces the coherence of the annotations but also illustrates how oral histories are structured and localized through specific lexical details. A complete breakdown of these entities by sentiment and semantic classification is provided in the appendix table~\ref{Entity}.

\subsection{Topic Modeling using BERTopic}

BERTopic\footnote{\url{https://github.com/MaartenGr/BERTopic}}, a topic modeling framework that integrates transformer based sentence embeddings with density based clustering and interpretable keyword extraction \cite{grootendorst2022bertopic}. We selected BERTopic for its ability to capture contextual semantics and uncover fine grained topics, which are essential when analyzing emotionally nuanced oral history narratives. Our data set comprises 92,191 sentences, each annotated with a sentiment label (Positive, Neutral, Negative) and a semantic category (classes 0-5). To ensure that topic modeling reflects sentiment specific themes, we applied BERTopic separately within each sentiment group. First, BERTopic generates contextual embeddings using sentence BERT, then applies HDBSCAN, a density based clustering algorithm, to group semantically similar sentences without predefining the number of topics. After clustering, class-based TF-IDF is applied to extract representative keywords for each topic by treating each cluster as a single class of text.

\begin{figure}[htbp]
    \centering
    \includegraphics[width=\textwidth]{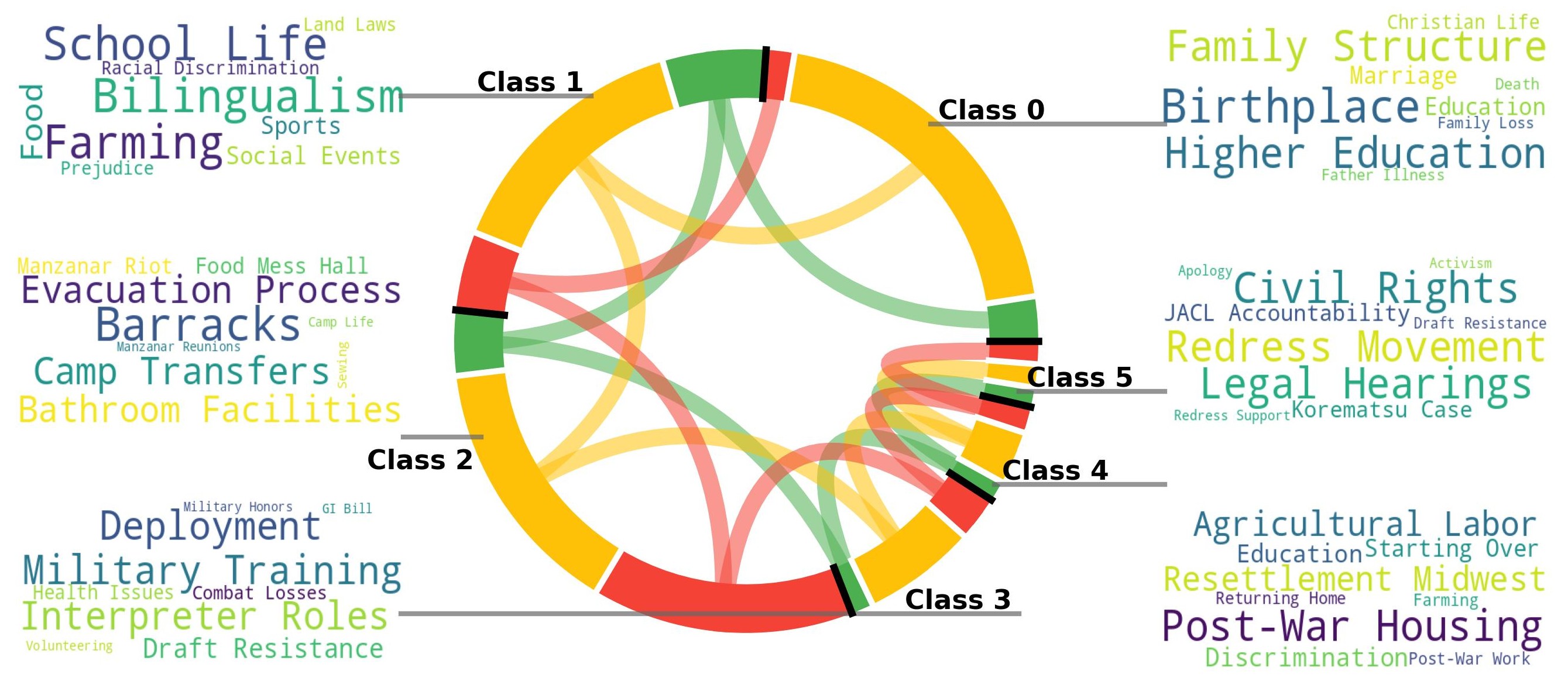}
    \vspace{-1.2em}
    \caption{
        Chord diagram of six semantic classes from Japanese-American incarceration narratives. 
        Arcs represent classes, colored by sentiment 
        (\textcolor{green!60!black}{Positive}, \textcolor{yellow!70!black}{Neutral}, \textcolor{red!70!black}{Negative}), 
        with chords linking shared sentiments across classes. 
        Word clouds show the top topics per class and sentiment.
    }
    \Description{Chord Diagram}
    \label{fig:chord_topics}
    \vspace{-0.3cm}
\end{figure}

To ensure topic quality and alignment with domain specific categories, we analyzed the top keywords from each cluster, assessed their frequency within each category, and manually interpreted and refined the resulting topics. We also extracted the top 10 keywords for each pair of sentiment category and manually assigned descriptive thematic labels. Figure~\ref{fig:chord_topics} illustrates a subset of the selected topics with word clouds and chord connections, the complete topic list is provided in Appendix Table~\ref{BERTopic}, reflects both the diversity and emotional framing of Japanese-American incarceration experiences. Neutral narratives focused on factual topics such as schooling, military service, and relocation. Positive narratives highlighted resilience, family life, achievements, and community involvement. In contrast, the negative narratives captured painful experiences, including FBI arrests, racial discrimination, illness, and harsh camp conditions. Importantly, several themes, such as education or camp life, appeared in sentiments but with markedly different emotional tones, underscoring the influence of sentiment on narrative framing. The full implementation pipeline and topic modeling results are available in public GitHub repository.

\section{Discussion}


\subsection{Research Question 1: How can we create a high quality, large-scale dataset for automatically understanding oral history?}

To create a high quality, large scale dataset for automatically understanding Japanese American oral history, it is essential to combine human expertise with scalable AI tools. The process should begin with a small, carefully selected set of interview transcripts that are manually annotated at the sentence level for both semantic categories (e.g. biographical details, incarceration experiences, redress movements) and sentiment (positive, neutral, negative). These annotations should be produced and validated by experts in the domain familiar with the cultural and historical context to preserve emotional nuance and historical accuracy. This gold-standard subset then serves as the basis for evaluating and guiding automated methods. Once this foundation is established, large scale annotation can be performed using prompt based LLMs. Practitioners should experiment with different prompt types, including zero-shot, few-shot, and RAG based approaches, while selecting the strategy that best balances accuracy, consistency, and efficiency for their specific task and model. In our case, concise few shot prompts with sentiment analysis, and few shot refined prompt for semantic classication worked best across models like ChatGPT, Llama, and Qwen, but others may find different prompt structures more effective depending on the content and domain. Regardless of the strategy, it is critical to ensure annotation quality through stability testing and downstream validation methods such as topic modeling or entity analysis. By combining rigorous human curation with adaptive LLM strategies, researchers can construct datasets that are not only scalable, but also culturally and ethically grounded.

\subsection{Research Question 2: What factors should be considered when selecting the most effective and efficient prompt strategy for annotating historically sensitive oral history data?}

While choosing the best prompt strategy for annotating historically rich oral histories, it is crucial to evaluate the overall efficiency of the approach balancing human effort, LLM inference cost, accuracy, and stability. Some strategies require more human involvement, such as curating examples or verifying model outputs, while others demand higher computational resources, especially when using longer prompts or retrieval-based methods. At the same time, models can vary in how consistently they perform on different inputs or runs, making stability a key concern. A truly efficient strategy is not just the one that gives the highest accuracy, it is the one that achieves reliable, high-quality results with minimal annotation fatigue, manageable API cost, and low output variance. Therefore, practitioners should compare prompt strategies through controlled experiments, assessing not only how well they classify categories and sentiment, but also how resource intensive and scalable they are in real-world settings. This multidimensional evaluation is essential for ensuring both practical feasibility and methodological rigor in large scale historical annotation process.

\subsection{Research Question 3: To what extent can large language models preserve contextual nuance and narrative integrity when applied to large-scale annotations of testimonies from marginalized communities?}

LLMs demonstrate strong consistency in sentiment analysis across architectures, suggesting that they can reliably capture general emotional tone. However, preserving contextual nuance and narrative integrity in semantic classification, especially within marginalized communities, is more challenging. In our study on Japanese-American oral histories, ChatGPT significantly outperformed Llama and Qwen in thematic annotation, highlighting that not all models equally internalize culturally embedded knowledge. This suggests that LLMs vary in their ability to preserve historical and narrative specificity, particularly when the domain context is implicit or culturally nuanced. In addition, subtle narrative shifts, such as layered memory, can be flattened or misclassified by models that lack exposure to such discourse. Therefore, to ensure that large-scale annotations truly reflect the depth and meaning of marginalized testimonies, it is important to go beyond just measuring the accuracy of the model. Researchers must also carefully check whether the model output preserves the original context and intent of the narratives, so that important cultural and historical details are not lost or misrepresented during automated processing.

\subsection{Research Question 4: What are the limitations and strengths of LLM-driven annotation pipelines in producing reliable and scalable semantic and sentiment labels for under described oral history collections?}

LLM driven annotation pipelines offer significant strengths in producing reliable and scalable sentiment and semantic labels for under described oral history collections. They enable large scale annotation with minimal manual effort, allowing massive sentences to be processed quickly and consistently. This scalability makes them especially valuable for archival projects that lack the resources for extensive human annotation. Furthermore, LLMs demonstrate strong performance in sentiment and semantic classification, consistently achieving high accuracy across different prompting strategies. Their ability to detect emotional tone in narrative rich testimonies makes them well suited for identifying affective patterns in historical discourse. However, these pipelines also have important limitations. While sentiment annotation is relatively robust, semantic labeling, particularly in culturally nuanced or historically complex narratives, still requires human in the loop oversight. LLMs struggle to capture subtle thematic distinctions without domain specific examples or few shot prompts. Their outputs are highly sensitive to prompt quality, often requiring iterative refinement informed by both human judgment and model feedback. In this way, the annotation process becomes a codependent effort, human annotators rely on LLMs for scale and assistive refinement, while the models depend on curated guidance to achieve interpretive accuracy. As such, fully automated annotation remains insufficient for high stakes oral history work, where preserving narrative integrity and contextual meaning is paramount.

\section{Conclusion and Future work}


Our study demonstrates how computational methods, when applied thoughtfully, can improve the accessibility, structure, and interpretability of large scale oral history archives. By integrating language models into the annotation of Japanese American incarceration narratives, we not only facilitate scalable processing of testimony rich data but also engage critically with the ethical and methodological challenges of preserving marginalized voices through automation. Importantly, this work goes beyond technical execution to foreground the questions of narrative fidelity, cultural context, and representational responsibility, highlighting the need for human-in-the-loop systems in sensitive historical domains. As digital archives continue to expand, our approach offers a replicable and adaptable framework for scholars, educators, and community institutions committed to inclusive memory work. Future research can build on this foundation by advancing culturally adaptive LLMs, exploring multilingual oral histories, and designing participatory annotation pipelines that further democratize historical interpretation. In addition, this framework opens new pathways for interdisciplinary collaboration, enabling historians, technologists, linguists, and community advocates to co-create tools that honor both data integrity and historical truth. Finally, by embedding structure and semantic depth in testimony rich archives, this work contributes to the long-term sustainability and discoverability of oral history collections, ensuring that they remain usable and meaningful for generations to come.

\subsection{Theoretical and Practical Implications}

This study contributes to both archival practice and digital humanities research. Our study offers a clear and replicable framework for using LLMs to annotate and analyze complex oral histories in a way that balances scale, accuracy, and respect for narrative integrity. Many archiving projects require texts to be classified into detailed categories. This work is often slow and relies on manual labor. Our project describes a framework that can automate the annotation process using open-source LLMs. Our study has many practical implications for the archiving community, researchers, and society through the creation of accessible annotated datasets and a scalable methodology for the preservation of oral history. The annotated large-scale dataset directly impacts the archiving community by providing labeled data that can be immediately utilized in various research projects. The comprehensive annotation of 92,191 sentences creates a valuable resource that other researchers can use to improve model performance or fine tune different language models specifically for oral history documents. Our data can become a foundation for future research in digital archiving by enabling the development of sophisticated tools for the analysis of personal narratives and historical testimonies. By applying sentiment and semantic analysis to oral history transcripts, this work enables the integration of thematically organized narratives into educational materials, making it easier to incorporate lived experiences into history books and curriculum, as McLellan emphasized \cite{mclellan1998case}. Beyond education, structured analysis of oral histories can inform public policy by surfacing community experiences and historically grounded insights that align with Hoffman’s call to connect policymaking with the lived realities of those affected \cite{hoffman2017practicing}.

These contributions signal a broader shift in how oral histories can be preserved, interpreted, and mobilized on a scale. As large annotated collections become more available, they open up new possibilities for interdisciplinary collaboration among historians, educators, technologists, and community organizations. This approach not only enhances archival workflows, but also empowers communities to reclaim and reinterpret their narratives through computational means. Through the showcase, this work demonstrates the thoughtful integration of human expertise, machine learning, and AI can help protect the integrity of lived experience while expanding its reach across research, education, and public discourse. It also bridges archival ethics, computational linguistics, and digital memory work by proposing an adaptable framework for ethically grounded AI integration in the digital humanities.



\subsection{Limitations}

This study has limitations. First, the evaluation and framework are based on a single dataset: the Japanese-American Incarceration Oral History (JAIOH) corpus. Although this data set is rich and thematically diverse, its historical and cultural specificity may limit the generalizability of the results. Further validation is needed to assess whether the proposed annotation approach performs equally well on other oral history collections with different narrative styles, linguistic features, or cultural contexts. Second, although the prompting strategies used in this study achieved strong results, particularly in sentiment and semantic classification, there is still significant room to improve the precision of sentiment and semantic annotation. The prompts were designed through careful experimentation, but more advanced strategies, including dynamic prompt tuning or hybrid human in the loop approaches, could further enhance performance. As prompt engineering continues to evolve, future work should explore systematic methods to refine instructions and better align model outputs with nuanced historical content.


\subsection{Future Work}

Future work will focus on expanding the applicability and robustness of the proposed annotation framework. First, to address the current limitation of the data set, we plan to evaluate the framework on additional oral history corpora from diverse cultural, linguistic, and historical backgrounds. This will allow us to test the adaptability of our methods and improve generalizability across underrepresented narratives. Second, there is substantial opportunity to refine the prompt design for improved semantic classification. Future efforts will explore automated prompt optimization techniques, retrieval-augmented prompting, and hybrid pipelines that incorporate human feedback to improve annotation precision. In addition, integrating domain-adaptive fine-tuning or a few-shot learning across tasks can further align LLM outputs with community-centered values and interpretive depth. Last but not least, we envision building interactive tools that allow historians, educators, and communities to explore annotated oral histories dynamically, supporting both scholarly research and public engagement with lived historical experiences.

\bibliographystyle{plain}
\bibliography{references}

\section{Appendices}



\subsection{Prompt Templates for Sentiment and Semantic Classification}
\label{sec:prompt_templates}
\subsubsection{Sentiment Classification:}\hfill

{\small
\begin{tcolorbox}[colback=gray!5, colframe=black, boxrule=1pt, rounded corners, title=\textbf{1. Foundational Prompt: Instruction + Sentiment list}, fonttitle=\bfseries]

\textbf{Prompt:}\\ 
\begin{tabularx}{\linewidth}{@{}lX@{}}
\textit{Instruction:} & \textless \textit{Instructions to the model}\textgreater \\
\textit{Sentiment List:} & \textless \textit{List of all the sentiments}\textgreater \\
\end{tabularx}

\textbf{Post:}\\
\textless \textit{The sentence that needs to be classified.}\textgreater 
\end{tcolorbox}
}

{\small
\begin{tcolorbox}[colback=gray!5, colframe=black, boxrule=1pt, rounded corners, title=\textbf{2. Structured Prompt: Instruction + Sentiment list + Sentiment Definitions}, fonttitle=\bfseries]

\textbf{Prompt:}\\ 
\begin{tabularx}{\linewidth}{@{}lX@{}}
\textit{Instruction:} & \textless \textit{Instructions to the model}\textgreater \\
\textit{Sentiment List:} & \textless \textit{List of all the sentiments}\textgreater \\
\textit{Sentiment Definitions:} & \textless \textit{Clear description of the sentiments}\textgreater \\
\end{tabularx}

\textbf{Post:}\\
\textless \textit{The sentence that needs to be classified.}\textgreater 
\end{tcolorbox}
}

{\small
\begin{tcolorbox}[colback=gray!5, colframe=black, boxrule=1pt, rounded corners, title=\textbf{3. Comprehensive Prompt: Instruction + Sentiment list + Sentiment Definitions +  Japanese American Incarceration Oral History Background}, fonttitle=\bfseries]

\textbf{Prompt:}\\ 
\begin{tabularx}{\linewidth}{@{}lX@{}}
\textit{Instruction:} & \textless \textit{Instructions to the model}\textgreater \\
\textit{Sentiment List:} & \textless \textit{List of all the sentiments}\textgreater \\
\textit{Sentiment Definitions:} & \textless \textit{Clear description of the sentiments}\textgreater \\
\textit{Background:} & \textless \textit{Background of Japanese American Incarceration Oral History}\textgreater \\
\end{tabularx}
\textbf{Post:}\\
\textless \textit{The sentence that needs to be classified.}\textgreater 
\end{tcolorbox}
}

{\small
\begin{tcolorbox}[colback=gray!5, colframe=black, boxrule=1pt, rounded corners, title=\textbf{4. Concise Prompt: Instruction + Concise Summary}, fonttitle=\bfseries]

\textbf{Prompt:}\\ 
\begin{tabularx}{\linewidth}{@{}lX@{}}
\textit{Instruction:} & \textless \textit{Instructions to the model}\textgreater \\
\textit{Concise Summary:} & \textless \textit{A concise summary of the classification task}\textgreater \\
\end{tabularx}

\textbf{Post:}\\
\textless \textit{The sentence that needs to be classified.}\textgreater 
\end{tcolorbox}
}

\subsubsection{Semantic Classification:}\hfill

{\small
\begin{tcolorbox}[colback=gray!5, colframe=black, boxrule=1pt, rounded corners, title=\textbf{1. Foundational Prompt: Instructions + Category list}, fonttitle=\bfseries]

\textbf{Prompt:}\\ 
\begin{tabularx}{\linewidth}{@{}lX@{}}
\textit{Instruction:} & \textless \textit{Instructions to the model}\textgreater \\
\textit{Category List:} & \textless \textit{List of all the categories}\textgreater \\
\end{tabularx}

\textbf{Post:}\\
\textless \textit{The sentence that needs to be classified.}\textgreater 
\end{tcolorbox}
}

{\small
\begin{tcolorbox}[colback=gray!5, colframe=black, boxrule=1pt, rounded corners, title=\textbf{2. Structured Prompt: Instruction + Category list + Category Definitions}, fonttitle=\bfseries]

\textbf{Prompt:}\\ 
\begin{tabularx}{\linewidth}{@{}lX@{}}
\textit{Instruction:} & \textless \textit{Instructions to the model}\textgreater \\
\textit{Category List:} & \textless \textit{List of all the categories}\textgreater \\
\textit{Category Definitions:} & \textless \textit{Clear description of the categories}\textgreater \\
\end{tabularx}

\textbf{Post:}\\
\textless \textit{The sentence that needs to be classified.}\textgreater 
\end{tcolorbox}
}

%



{\small
\begin{tcolorbox}[colback=gray!5, colframe=black, boxrule=1pt, rounded corners, title=\textbf{3. Comprehensive Prompt: Instructions + Category list + Category definitions + Background information of Japanese American Incarceration + Keywords}, fonttitle=\bfseries]

\textbf{Prompt:}\\ 
\begin{tabularx}{\linewidth}{@{}lX@{}}
\textit{Instruction:} & \textless \textit{Instructions to the model}\textgreater \\
\textit{Category List:} & \textless \textit{List of all the categories}\textgreater \\
\textit{Category Definitions:} & \textless \textit{Clear description of the categories}\textgreater \\
\textit{Background:} & \textless \textit{Background of Japanese American Incarceration Oral History}\textgreater \\
\textit{Keywords:} & \textless \textit{Keywords extracted for each category}\textgreater \\
\end{tabularx}

\textbf{Post:}\\
\textless \textit{The sentence that needs to be classified.}\textgreater 
\end{tcolorbox}
}

{\small
\begin{tcolorbox}[colback=gray!5, colframe=black, boxrule=1pt, rounded corners, title=\textbf{4. Refined Prompt: Instructions + Category list + Background information of Japanese American Incarceration + Keywords + ChatGPT Definitions}, fonttitle=\bfseries]

\textbf{Prompt:}\\ 
\begin{tabularx}{\linewidth}{@{}lX@{}}
\textit{Instruction:} & \textless \textit{Instructions to the model}\textgreater \\
\textit{Category List:} & \textless \textit{List of all the categories}\textgreater \\
\textit{Background:} & \textless \textit{Background of Japanese American Incarceration}\textgreater \\
\textit{Keywords:} & \textless \textit{Keywords extracted for each category}\textgreater \\
\textit{ChatGPT Definitions:} & \textless \textit{Clear description of each category given by ChatGPT}\textgreater \\
\end{tabularx}

\textbf{Post:}\\
\textless \textit{The sentence that needs to be classified.}\textgreater 
\end{tcolorbox}
}

{\small
\begin{tcolorbox}[colback=gray!5, colframe=black, boxrule=1pt, rounded corners, title=\textbf{5. Concise Prompt: Instruction + Concise Summary}, fonttitle=\bfseries]

\textbf{Prompt:}\\
\begin{tabularx}{\linewidth}{@{}lX@{}}
\textit{Instruction:} & \textless \textit{Instructions to the model}\textgreater \\
\textit{Concise Summary:} & \textless \textit{A concise summary of the classification task}\textgreater \\
\end{tabularx}

\textbf{Post:}\\
\textless \textit{The sentence that needs to be classified.}\textgreater 
\end{tcolorbox}
}

More detailed prompts can be accessed at the \href{https://github.com/kc6699c/LLM4OralHistoryAnalysis}{Repository}.

\newpage

\subsection{Top Entities Across Different Sentiment and Semantic Classes}

\small
\begin{longtable}{|c|l|p{12.6cm}|}
\hline
\textbf{Sentiment} & \textbf{Category} & \textbf{Entities} \\
\hline
\endfirsthead
\hline

\multirow{6}{*}{Neutral}
 & 0 & \textit{FAC} — Terminal Island, Seabrook, Bainbridge Island, Fort Lupton, Main Street, Jackson Street, Broadway, Weller Street, Angel Island, Little Tokyo Towers, \textit{GPE} — Japan, Seattle, Hawaii, California, the United States, San Francisco, Portland, Hiroshima, Tokyo, \textit{LANGUAGE} — English, Spanish, Filipino, Japanese, \textit{LOC} — Angel Island, Bay Area, East Coast, South America, Southern California, Hood River, \textit{NORP} — Japanese, Buddhist, American, Christian, Catholic, Caucasian, Chinese, Japanese American, Baptist \textit{ORG} — UCLA, USC, University of Washington, Sansei, \textit{QUANTITY} — ten acres, about five foot.
   \\ \cline{2-3}
 & 1 & \textit{EVENT:} World War II, New Year, \textit{FAC:} Main Street, Pearl Harbor, Terminal Island, Broadway, Jackson Street, First Street, \textit{GPE:} Japan, Seattle, California, Los Angeles, San Francisco, Portland, Pacoima, \textit{ORG:} UCLA, the University of Washington, Gardena, Lincoln, Bainbridge, Methodist, \textit{PERSON:} Dad, Mom, \textit{LANGUAGE:} English, Filipino, Japanese, Spanish \textit{LAW:} Constitution, the Exclusion Act, the Pledge of Allegiance, the Railroad Act, the Alien Land Law, \textit{LOC:} Terminal Island, Hood River, the West Coast, Midwest, Bainbridge Island, the International District, \textit{NORP:} Japanese, American, Chinese, Caucasian, Buddhist, Japanese Americans, \textit{QUANTITY:} five acres, twenty acres, forty acres  \\ \cline{2-3}
 & 2 & \textit{EVENT:}  Camp, World War II, \textit{FAC:} Pearl Harbor, Tule Lake, Camp Savage, Camp Shelby, Terminal Island, Block 1, Barrack 8, Block 29, Block 22, Block 28, Block 32, Block 5, \textit{GPE:} Manzanar, Topaz, Japan, California, Santa Anita, Seattle, \textit{ORG:} Heart Mountain, FBI, WRA, Amache, Santa Fe, Puyallup, Minidoka, Jerome, Rohwer, Gila, \textit{DATE:} May, 1942 \textit{LANGUAGE:} English, Japanese, Spanish,  \textit{LOC:} Crystal City, the West Coast, Delta, Terminal Island, Hood River, \textit{NORP:} Japanese, American, Caucasian, Buddhist, Japanese Americans, German  \\ \cline{2-3}
 & 3 & \textit{EVENT:} World War II, World War I, the Korean War, Platoon, \textit{FAC:} Fort Snelling, Camp Shelby, Camp Savage, Fort Lewis, Camp Blanding, Fort Ord, Navy, \textit{GPE:} Japan,  Hawaii, the United States, Washington, Tokyo, \textit{ORG:} 442nd, MIS, Army, GI, the Air Force, \textit{PERSON:} George, Harry, Joe, Monterey, \textit{WORK\_OF\_ART:} Lost Battalion, the Battle of the Lost Battalion, \textit{LANGUAGE:} English, Japanese, Chinese, Filipino, \textit{LAW:} Constitution, Section 6J, \textit{LOC:} Europe, Pacific, the South Pacific, Far East, the West Coast, the Pacific,\textit{ NORP:} Japanese, American, Japanese American, Germans (13) \\ 
 & 4 & EVENT: World War II, New Year' s Eve, Olympics, \textit{FAC:} Seabrook, First Street, Fort Dix, Seabrook Farms, Tokyo Bay, Tule Lake, Twin Falls \textit{GPE:} Japan, Seattle, Chicago, California, Portland, Los Angeles, \textit{ORG:} Bainbridge, GI, Caldwell, the University of Washington, Gardena, Kyushu, \textit{PERSON:} Dad, \textit{WORK\_OF\_ART:} PhD, Alameda Naval Air Station, \textit{DATE:} 1945, 1946, 1948, 1947, \textit{LANGUAGE:} English, \textit{LAW:} the McCarran Act, \textit{LOC:} the West Coast, East, Midwest, the East Coast, Europe, Hood River, \textit{NORP:} Japanese, American, Japanese Americans, Chinese, Buddhist, Americans   \\ \cline{2-3}
 & 5 & \textit{EVENT:} World War II, Remembrance Day, the Civil War (1), the Vietnam War, \textit{FAC:} Capitol, Pearl Harbor, Boalt Hall, Isamu Noguchi, Little Tokyo, the Golden Gate JACL, JACL, \textit{GPE:} Seattle, Washington, Los Angeles, San Francisco, Manzanar, \textit{ORG:} Congress, NCJAR, Heart Mountain, ACLU, American Citizens League, Commission, \textit{PERSON:} Inouye, Reagan, Mike Masaoka, Matsunaga, Mineta, \textit{WORK\_OF\_ART:} LTPRO, Little Tokyo People' s Rights Organization, Stray Cats of Manzanar, the Issei History Project, \textit{DATE:} 1988, 1970, ' 80s, ' 70s, 1980, the Day of Remembrance, \textit{LAW:} the Civil Liberties Act, Constitutional, Civil Liberties Act, the Internal Security Act, the McCarran Act, LOC: the West Coast, the Bay Area, Mountain View, Bay Area, Southern California, South, \textit{MONEY:} 20,000, twenty thousand dollars, 25,000, \textit{NORP:} Japanese, American, Japanese Americans, Asian, Japanese American, Americans
\\ \hline
\multirow{6}{*}{Positive}
 & 0 & \textit{EVENT:} Olympic, New Year, \textit{FAC:} Broadway, Boyle Heights, Buchanan Street, Crowley Lake, Post Street, the Constitution Hall, \textit{GPE:} Japan, America, Hawaii, San Francisco, Chicago, Portland, \textit{ORG:} UCLA, Methodist, the Buddhist Church, Gardena, Lincoln, Waseda University, \textit{PERSON:} Dad, Grandma, Mary, Mom, Paul, \textit{WORK\_OF\_ART:} PhD, the Pledge of Allegiance, \textit{LANGUAGE:} English, Japanese, \textit{LOC:} the Bay Area, East Bay, Okudas, Sasaki, Tabernacle, the East Bay Japanese Camera Club, \textit{NORP:} Japanese, Buddhist, American, Christian, Japanese American, Baptist
 \\ 
 & 1 & \textit{EVENT:} \textit{EVENT:} New Year, the Rose Festival, \textit{FAC:} Jackson Street, Broadway, Collins Playfield, Little Tokyo, Main Street, the Japanese Hall, \textit{GPE:} Japan, Seattle, Portland, Los Angeles, Tokyo, Hawaii, \textit{ORG:} Methodist, Gardena, Lincoln High School, YMCA, Bainbridge, the Boy Scouts, \textit{PERSON:} Dad, Mom, Shigios, \textit{WORK\_OF\_ART:} the Wapato Nippons, PhD, \textit{LANGUAGE:} English, Filipino, Japanese, \textit{LAW:} Constitution, the "American Pickers", \textit{LOC:} the West Coast, Terminal Island, Mercer Island, East, Russian River, the International District, \textit{NORP:} Japanese, American, Buddhist, Caucasian, Japanese American, Japanese Americans
 \\ \cline{2-3}
 & 2 & 
\textit{EVENT:} New Year's, New Year's Eve, World War II, \textit{FAC:} Tule Lake Camp, Camp Savage, Camp 1, Camp 2, Block 1, Block 5, Block 34, the Pomona Hotel, \textit{GPE:} Manzanar, Topaz, California, Seattle, Santa Anita, Minidoka, Jerome, Gila River, Amache, \textit{ORG:} Heart Mountain Relocation Center, Quakers, Maryknoll, Puyallup Assembly Center, Santa Fe Internment Camp, \textit{PERSON:} Jerome, Mary, Sweetheart of Minidoka, \textit{WORK\_OF\_ART:} Camp Harmony, Sweetheart of Minidoka, The No Name Team, \textit{DATE:} Christmas, the summer, \textit{LANGUAGE:} English, Filipino, Spanish, Tagalog, \textit{LAW:} Barrack 10, \textit{LOC:} Crystal City, Heart Mountain, Gila River, the Colorado River, Bainbridge Island, \textit{NORP:} Japanese, American, Buddhist, Caucasian, Japanese Americans, Christian
 \\  \cline{2-3}
 & 3 &\textit{EVENT:} World War II, World War, Memorial Day, Labor Day, VE Day, \textit{FAC:} Fort Snelling, Camp Savage, Fort Belvoir, Camp Blanding, Buckley Air Force Base, Fort Carson, \textit{GPE:} United States, Japan, America, Italy, Hawaii, \textit{ORG:} 442nd Regimental Combat Team, GI, Military Intelligence Service, U.S. Army, USO, ROTC, United States Army, \textit{PERSON:} Dan Inouye, Inouye, Ben Kuroki, \textit{WORK\_OF\_ART:} PhD, Lost Battalion, Call Yoshinaga, Omaha, the Silver Star, \textit{DATE:} two years, today, four years, every day, three years, one day, \textit{LANGUAGE:} English, Spanish, \textit{LAW:} Constitution, G2, \textit{LOC:} Europe, Pacific, North Hollywood, the Far East, the Gothic Line, the West Coast, \textit{NORP:} Japanese, American, Americans, Japanese Americans, Japanese American, Caucasian
  \\ \cline{2-3}
 & 4 &
\textit{EVENT:} World War II, the Korean War, the Rose Festival Parade, the Second World War, \textit{FAC:} Bainbridge Gardens, Jackson Street, Pearl Harbor, Seabrook Farms, Jackson Cafe, \textit{GPE:} Japan, Seattle, Chicago, California, Portland, the United States, \textit{ORG:} GI, WRA, Heart Mountain, UCLA, the Red Cross, the University of Washington, \textit{PERSON:}  Mom, \textit{WORK\_OF\_ART:} PhD, \textit{DATE:} 1946, 1947, 1953, \textit{LANGUAGE:} English, \textit{LOC:} the West Coast, Crystal City, Hood River, Midwest, the East Coast, the Pacific Coast, \textit{NORP:} Japanese, American, Americans, Japanese Americans, Japanese American, Canadian
\\ \cline{2-3}
 & 5 & 
\textit{EVENT:} World War II, Title II, Vietnam War, the War on Poverty, the World War II, \textit{FAC:} Tule Lake, Bainbridge Island, Pearl Harbor, Point Arena, San Francisco State, the Issei History Project, \textit{GPE:} Washington, Seattle, Manzanar, the United States, \textit{ORG:} JACL, 442nd Regimental Combat Team, Congress, American Citizens League, Heart Mountain, ORA, the Civil Rights Movement, \textit{PERSON:} Inouye, Gordon Hirabayashi, Ronald Reagan, Norman Mineta, Fred Korematsu, Edison Uno, \textit{WORK\_OF\_ART:} Save Little Tokyo, Better Americans in a Greater America, Leaving Our Island, Leaving the Island, The Issei Pioneers, The Nisei Something, \textit{DATE:} 1988, \textit{LANGUAGE:} English, Filipino, \textit{LAW:} Constitution, the Fair Play Committee, the Civil Liberties Act, the Report of the Commission on Wartime Relocation and Internment of Civilians, the First Compensation Act, the Nationality Act, \textit{LOC:} the Bay Area, Bay Area, Delta, Europe, the West Coast, Leavenworth, \textit{NORP:} Japanese, American, Japanese Americans, Asian, Americans, Japanese American
 \\ \hline
\multirow{6}{*}{Negative} 
 & 0 &
\textit{EVENT:} New Year, World War II, \textit{FAC:} Sand Creek, Terminal Island, Wapato, Broadway High School, Nishimoto Trading, Port Moody, \textit{GPE:} Japan, America, the United States, Hawaii, Hiroshima, Tokyo, \textit{ORG:} Methodist, aNisei, \textit{PERSON:} Dad, Mom, Frank, Joe, Grandma, Clarence, Papa, Takeshi, \textit{LANGUAGE:} English, \textit{LOC:} Crystal City, Northern Montana, Pacific, South America, the Pacific Ocean, the West Coast, \textit{NORP:} Japanese, American, Buddhist, Christian, Asian, Caucasian
  \\ 
 & 1 & 
\textit{EVENT:} World War II, World War, New Year's, the American Revolution, the Nisei reputation, \textit{FAC:} Pearl Harbor, Terminal Island, Main Street, Broadway, Lincoln Park, Sixteenth Avenue, \textit{GPE:} Japan, California, Seattle, America, Hawaii, San Francisco, \textit{ORG:} UCLA, the University of Washington, Stanford, aNisei, the Isseis were, \textit{PERSON:} Dad, Mom, Jap, Japs, \textit{WORK\_OF\_ART:} No Japs, No Japs Allowed, No Japs Wanted, Kill the Jap, Yellow Peril, \textit{LANGUAGE:} English, Filipino, Spanish, \textit{LAW:} the Immigration Act, the Alien Exclusion Act, the Exclusion Act, Kejima-san, Section 4, the National Origins Act, \textit{LOC:} the West Coast, Hood River, Terminal Island, South, Asia, West Coast, \textit{NORP:} Japanese, American, Chinese, Caucasian, Asian, Americans
  \\ \cline{2-3}
 & 2 & 
\textit{EVENT:} World War II, World War, World War I, the Second War, \textit{FAC:} Pearl Harbor, Tule Lake, Terminal Island, Camp Shelby, Santa Anita, Bainbridge Island, Block 5, Block 4, \textit{GPE:} Japan, Manzanar, California, Seattle, \textit{ORG:} FBI, Heart Mountain, WRA, Bainbridge, Puyallup, Santa Fe, \textit{PERSON:} Dad, \textit{WORK\_OF\_ART:} Camp Harmony, No Japs, No Japs Allowed, Bomb the Jap, Oneesan, \textit{DATE:} December 7th, \textit{LANGUAGE:} English, Filipino, Japanese, Latin, \textit{LAW:} Constitution, the Bill of Rights, the Fair Play Committee, Section 4, Block 36, Camp 1, \textit{LOC:} the West Coast, Terminal Island, Crystal City, Heart Mountain
 \\  \cline{2-3}
 & 3 & 
\textit{EVENT:} World War II, the Korean War, the Pacific War, Memorial Day, the Second World War, the World War I days, \textit{FAC:} Camp Savage, Camp Shelby, Camp Robinson, Fort Lewis, Fort Snelling, Tule Lake, \textit{GPE:} Japan, the United States, America, Hawaii, France, \textit{ORG:} 442nd Regimental Combat Team, Navy, BAR, Heart Mountain, \textit{WORK\_OF\_ART:} Lost Battalion, Gee, Dad, the Battle of the Bulge, the Congressional Medal of Honor, \textit{LANGUAGE:} English, Filipino, \textit{LAW:} Constitution, Bill of Rights, the Foreign Wars, the Selective Service Act, \textit{LOC:} Pacific, Europe, Tule Lake, the South Pacific, Heart Mountain, the West Coast, \textit{NORP:} Japanese, American, Japanese Americans, Americans, Caucasian, Germans
 \\ \cline{2-3}
 & 4 & 
\textit{EVENT:} World War II, World War, New Year's, the Great Depression, the War Measures Act, \textit{FAC:} Seabrook, Bainbridge Island, Apricot District, Keio, Madison Avenue, Pearl Harbor, \textit{GPE:} Japan, California, Seattle, Hiroshima, Chicago, the United States, \textit{ORG:} Little Tokyo, GI, WRA, Firestone, the Hood River American Legion, \textit{PERSON:} Dad, MacArthur, \textit{WORK\_OF\_ART:} No Japs Allowed, No Japs, No Japs Wanted, Richard, \textit{DATE:} 1946, 1950, \textit{LANGUAGE:} English, Filipino, \textit{LAW:} the Immigration Act, \textit{LOC:} the West Coast, Hood River, Pacific, Crystal City, McNeil Island, the Bay Area, \textit{NORP:} Japanese, American, Japanese Americans, Americans, Caucasian, Asian
\\ \cline{2-3}
 & 5 & 
\textit{EVENT:} World War II, Holocaust, \textit{FAC:} Pearl Harbor, Golden Gate, Little Tokyo, Tule Lake, the Hilton Hotel, the Hokoku Seinendan, \textit{GPE:} the United States, Japan, America, California, San Francisco, \textit{ORG:} JACL, Congress, the Supreme Court, American Citizens League, the Civil Rights Movement, Heart Mountain, \textit{WORK\_OF\_ART:} Stereotypes \& Admonitions, From Relocation to Segregation, the Korematsu decision, \textit{DATE:} 1978, 1988, \textit{LAW:} Constitution, the Bill of Rights, the Civil Liberties Act, The Evacuation Claims Act, the Internal Security Act, the McCarran Act, \textit{LOC:} the West Coast, Crystal City, the International District, Europe, Bainbridge Island, Northern California, \textit{NORP:} Japanese, American, Japanese Americans, Americans, Japanese American, Asian
 \\ 
\hline
\caption{Entities extracted from the annotated oral history dataset, categorized by semantic type and sentiment label.}
\label{Entity}
\end{longtable}

\subsection{Major Topics Across Different Sentiment and Semantic Classes}

\begin{longtable}{|p{1.35cm}|p{1.25cm}|p{13cm}|} 
\hline
\textbf{Sentiment} & \textbf{Category} & \textbf{Topics} \\
\hline
\endfirsthead
\hline
\multirow{6}{*}{Neutral}
 & 0 & Biographical Information, Birthplace and Region, Childhood Age, Education and Language, Family Structure, Higher Education, Religious Affiliation – Christianity/Buddhism, Siblings, Marriage   \\ \cline{2-3}
 & 1 & Education and School Life, Daily Commute, Bilingualism, Ethnic Neighborhood, Housing and Property, Farming and Agricultural Labor, Religious Services and Practices, Stores and Commercial Areas, Fruit Farming  \\ \cline{2-3}
 & 2 & Incarceration Experience, Living Quarters, Barrack Assignments, Transportation to Camps, Camp Locations, Memories of Manzanar, Camp Transfers \& Releases, Barrack Layout \& Structure, Evacuation Process, Residential Blocks \\ \cline{2-3}
 & 3 & Military Rank and Structure, 442nd Regiment and Units, Draft and Enlistment, Military Service Career, Training and Deployment, War Service Reflections, Military Bases and Camps, Interpreter and Translation Roles, Military Discharge, Intelligence and Counterintelligence   \\ \cline{2-3}
 & 4 & Incarceration Experience, Post-War Housing and Relocation, Return to Washington Region, Returning Home, Post-War Timeline, Resettlement in Midwest Cities, War's End and Aftermath, Agricultural Labor, Japan Post-War and Occupation, Pacific Northwest Return  \\ \cline{2-3}
 & 5 & Redress Movement, JACL and Organizational Involvement, Civil Rights and Activism, Legal Hearings and Testimonies, Legislative Action and Compensation, Citizens Leagues and Legal Advocacy, Political Figures and Supporters, Landmark Legal Cases, Historical Internment Context, Compensation Amounts \\ \hline
\multirow{6}{*}{Positive}
 & 0 & Marriage and Relationships, Women and Family Roles, Educational and Career Path (Male), Christian Religious Life, Buddhist Religious Involvement, Female Identity and Heritage, Male Identity and Community, Educational Achievement, Maternal Characteristics, Japanese Language and Schooling \\ \cline{2-3}
 & 1 & Japanese Language and Schooling, Academic Achievement, Food and Dining, Social Events and Dances, Christian Religious Life, Farming and Produce, Sports and Athletics, Fishing Activities, Japanese Cooking and Cuisine, Buddhist Religious Involvement \\ \cline{2-3}
 & 2 & General Camp Life, Food and Dining, Social Events and Dances, Manzanar and Reunions, Sports and Athletics, Sewing and Clothing, Farming and Produce, Military Service and Citizenship, Music and Camp Bands, Christian Religious Life\\ \cline{2-3}
 & 3 & Military Service and Citizenship, 442nd Regiment and European Deployment, GI Bill and Education Benefits, Volunteering and Bravery, Military Medals and Honors, Japanese American Identity (Male), Broader Japanese American Community, Camp Life, Japanese Language and Schooling, Academic Achievement \\ \cline{2-3}
 & 4 & Returning Home and Community Reception, Camp Life, Military Service and Citizenship, Academic Achievement, Christian Religious Life, Post-War Work and Relocation, Store Life and Public Spaces, Family in Japan, Farming and Produce, Post-War Employment and Reputation \\ \cline{2-3}
 & 5 & Redress and Public Support, Apology and Compensation, JACL and Civil Rights Organizations, Japanese American Community Identity, Civil and Human Rights Advocacy, Draft Resisters and Moral Stance, Military Service and Citizenship, Legal Defense and Vindication, NCRR and Activist Legacy, Camp Life \\ 
 \hline
\multirow{6}{*}{Negative} 
 & 0 & Death and Loss(Female), Death of Siblings, Family and Siblings, Male Family Deaths, Name Pronunciation and Change, Father’s Death and Illness, Japanese Identity and Return, Employment and Career (Male), Relationship with Father/Stepfather, Loss of Relatives in Japan \\ \cline{2-3}
 & 1 & Racial Discrimination and Segregation, Anti-Japanese Slurs and Signs, School and English Language Learning, Japanese Schooling and Activities, Farming and Crop Work, Employment and Education (Male), Cultural Identity and Heritage, Female Identity and Return to Japan, Prejudice Against Japanese Americans, Land Ownership and Alien Laws \\ \cline{2-3}
 & 2 & Bathroom and Water Facilities, Food and Mess Halls, Train Travel and Shades, Pearl Harbor and Attack News, Tule Lake and Segregation, FBI Arrests and Detainment, Manzanar Riot and Camp Transport, Evacuation Process and Uncertainty, Barracks and Living Conditions, Packing \\ \cline{2-3}
 & 3 & Draft and Enlistment Process, Military Service Decisions, 442nd and 100th Battalion Experiences, Loyalty and Draft Resistance, Health and Camp Medical Issues, Military Rank and Leadership, Citizenship and Renunciation, Loyalty Questionnaire, Japanese Identity and Military Role, Combat and Battalion Losses \\ \cline{2-3}
 & 4 & Racial Discrimination and Segregation, Returning and Starting Over, Return from Japan Post-War, Farming and Agricultural Labor, Anti-Japanese Slurs and Public Sentiment, Education and Graduation, West Coast Restrictions, Camp Life and Individual Stories, Evacuation from Pacific Northwest, Japanese Identity and Military Involvement \\ \cline{2-3}
 & 5 & JACL and Leadership Accountability, Redress Movement and Public Opinion, Civil Rights and Constitutional Cases, Internment Experience Reflections, Japanese American Legal Advocacy, Racial Discrimination and Segregation, Korematsu Case and Related Trials, Formal Apologies and Acknowledgments, Draft Resistance and Patriotism, Compensation and Monetary Settlement \\ \hline
\caption{Topic distribution across semantic categories and sentiment classes}
\label{BERTopic}
\end{longtable}

\end{document}